\documentclass[journal ]{new-aiaa}
\usepackage[utf8]{inputenc}
\usepackage{textcomp}

\usepackage{graphicx}
\usepackage{amsmath}
\usepackage[version=4]{mhchem}
\usepackage{siunitx}
\usepackage{longtable,tabularx}
\usepackage{booktabs}
\usepackage{algorithm} 
\usepackage{algpseudocode} 
\usepackage{caption}
\usepackage{subcaption}
\usepackage{multirow}

\title{Deep-Dispatch: A Deep Reinforcement Learning-Based Vehicle Dispatch Algorithm for Advanced Air Mobility}

\author{Elaheh Sabziyan Varnousfaderani\footnote{Graduate Research Assistant, College of Aeronautics and Engineering, esabziya@kent.edu}, Syed A. M. Shihab \footnote{Assistant Professor, College of Aeronautics and Engineering, AIAA Member, sshihab@kent.edu}, and Esrat F. Dulia\footnote{Graduate Research Assistant, College of Aeronautics and Engineering, edulia@kent.edu}}
\affil{Kent State University, Kent, OH, USA 44242}

\begin{document}

\maketitle

\begin{abstract}
\label{abstract}


Near future air taxi operations with electric vertical take-off and landing (eVTOL) aircraft will be constrained by the need for frequent recharging of eVTOLs, limited takeoff and landing pads in vertiports, and subject to time-varying demand and electricity prices, making the eVTOL dispatch problem unique and particularly challenging to solve. Previously, we have developed optimization models to address this problem. Such optimization models however suffer from prohibitively high computational run times when the scale of the problem increases, making them less practical for real world implementation. To overcome this issue, we have developed two deep reinforcement learning-based eVTOL dispatch algorithms, namely single-agent and multi-agent deep Q-learning eVTOL dispatch algorithms, where the objective is to maximize operating profit. An eVTOL-based passenger transportation simulation environment was built to assess the performance of our algorithms across $36$ numerical cases with varying number of eVTOLs, vertiports, and demand. The results indicate that the multi-agent eVTOL dispatch algorithm can closely approximate the optimal dispatch policy with significantly less computational expenses compared to the benchmark optimization model. The multi-agent algorithm was found to outperform the single-agent counterpart with respect to both profits generated and training time.

\end{abstract}

\section*{Nomenclature}

{\renewcommand\arraystretch{1.0}
\noindent\begin{longtable*}{@{}l @{\quad=\quad} l@{}}
$N$  & number of eVTOLs in the AAM air taxi operator's fleet \\
$M$ &    number of vertiports in the AAM air taxi operator's network\\
$n$& number of time steps in the dispatch horizon or episode\\
$i$& index for eVTOLs: $i \in {1, ..., N}$\\
$j, k$& index for vertiports: $j,k \in {1, ..., M}$\\
$t$ & time step: $t \in \{1,..., n\}$\\
$\tau$ & time interval of a time step\\
$T$ & total time interval of a dispatch episode\\
$R_{f}$   & set of feasible routes \\
$V$   & set of vertiports: $V = \{1,...,M\}$ \\
$v$   & limited number of takeoff and landing pads in each vertiport \\
$D(t)$ & vector of total passenger demand for each route at time step $t$ \\
$B_{i}(t)$  & battery level of eVTOL $i$ at time step $t$\\
$B_{max}$ & battery capacity of eVTOLs \\
$\eta$  & battery discharging rate \\
$L_{i}(t)$  & location of eVTOL $i$ at time step $t$ \\
$l_{jk}$  & length of the route between vertiports $j$ and $k$ \\
$l_{max}$  & maximum length of route out of all routes \\
$E(t)$   & electricity price at time step $t$ \\
$E_{max}$ &    maximum electricity price over an episode \\
$w_{jk}(t)$  & number of passengers transported by an eVTOL from vertiport $j$ to vertiport $k$ at time step $t$\\
$ \Tilde{t}_{jk}$ & the time takes for an eVTOL to travel from vertiport $j$ to vertiport $k$ \\ 
$\rho$  & trip fare per passenger per mile\\
$\rho_{o}$  & operating cost per mile per available seat\\
$s(t)$  & state of the single agent at time step $t$\\
$\hat{s}_{e}(t)$ & state of all eVTOLs at time step $t$\\
$\hat{s}_{i}(t)$  & state of eVTOL $i$ \\
${s}^{\prime}_{i}(t)$  & state of agent $i$ at time step $t$\\
$A(t)$  & set of available actions for single agent \\
$A^{\prime}(t)$  & set of available actions for multi agents \\
$\hat{a}_{i}(t)$  & action taken by the single agent for eVTOL $i$ at time step $t$ \\
${a}^{\prime}_{i}(t)$  & action taken by agent $i$ at time step $t$ \\
$\hat{r}_{i}(t)$  & reward received by the single agent with taking action $\hat{a}_{i}(t)$ for eVTOL $i$ at time step $t$\\
${r}^{\prime}_{i}(t)$  & reward received by agent $i$ by taking action ${a}^{\prime}_{i}(t)$ at time step $t$\\
$R$  & total cumulative reward received by the single agent for dispatching a fleet of $N$ eVTOLs at the end of the episode\\
${R}^{\prime}$  & total cumulative reward received by $N$ agents at the end of the episode\\
\end{longtable*}}

\section{Introduction}
\label{Introduction}
\subsection{Advanced Air Mobility}
 
Traffic congestion has become a serious problem in many urban cities. Many commuters spend a huge amount of time and money in traffic in the United States. In 2022, drivers lost a staggering 4.8 billion hours to traffic jams, which cost them a total of \$81 billion across the United States \cite{Traffic}. In addition to the traffic congestion in urban cities, suburban and rural regions also suffer from limited transportation options and extended travel times. This highlights the urgent need for an innovative transportation solution that can mitigate the traffic congestion problem. 

In response to these challenges, the concept of Advanced Air Mobility (AAM) has shown promise as an affordable, faster, and accessible mode of intracity and intercity air travel. AAM envisions a safe, automated, and efficient air transportation system in the low altitude airspace for passengers and cargo, connecting urban, surburan, and rural areas, even those that are hard to reach \cite{northeast2020advanced}. Market studies commissioned by NASA project that by 2030 there will be up to 750 million flights per year for passenger transportation services and up to 500 million flights per year for package delivery services \cite{NASA1}. To bring AAM to life, various organizations --- such as NASA, Volocopter, Uber, Airbus, Aurora Flight Sciences and Joby Aviation --- are striving to develop low altitude aircraft, and tackle market barriers and operational issues related to implementation of AAM \cite{NASA,Volocopter,Uber,Airbus,Aurora,JobyAviation}. AAM is conceptualized to employ various types of electric vertical takeoff and landing (eVTOL) aircraft to transport passengers and cargo between vertiports. eVTOLs can takeoff, land, and recharge at vertiports. These vertiports can be situated on top of existing buildings or constructed as standalone structures \cite{NASA1}. In addition, existing airports for traditional aircraft can be expanded to act as vertiports for eVTOLs.

\begin{figure}[ht]
\centering
\includegraphics[width=16.5 cm, height = 8.5 cm]{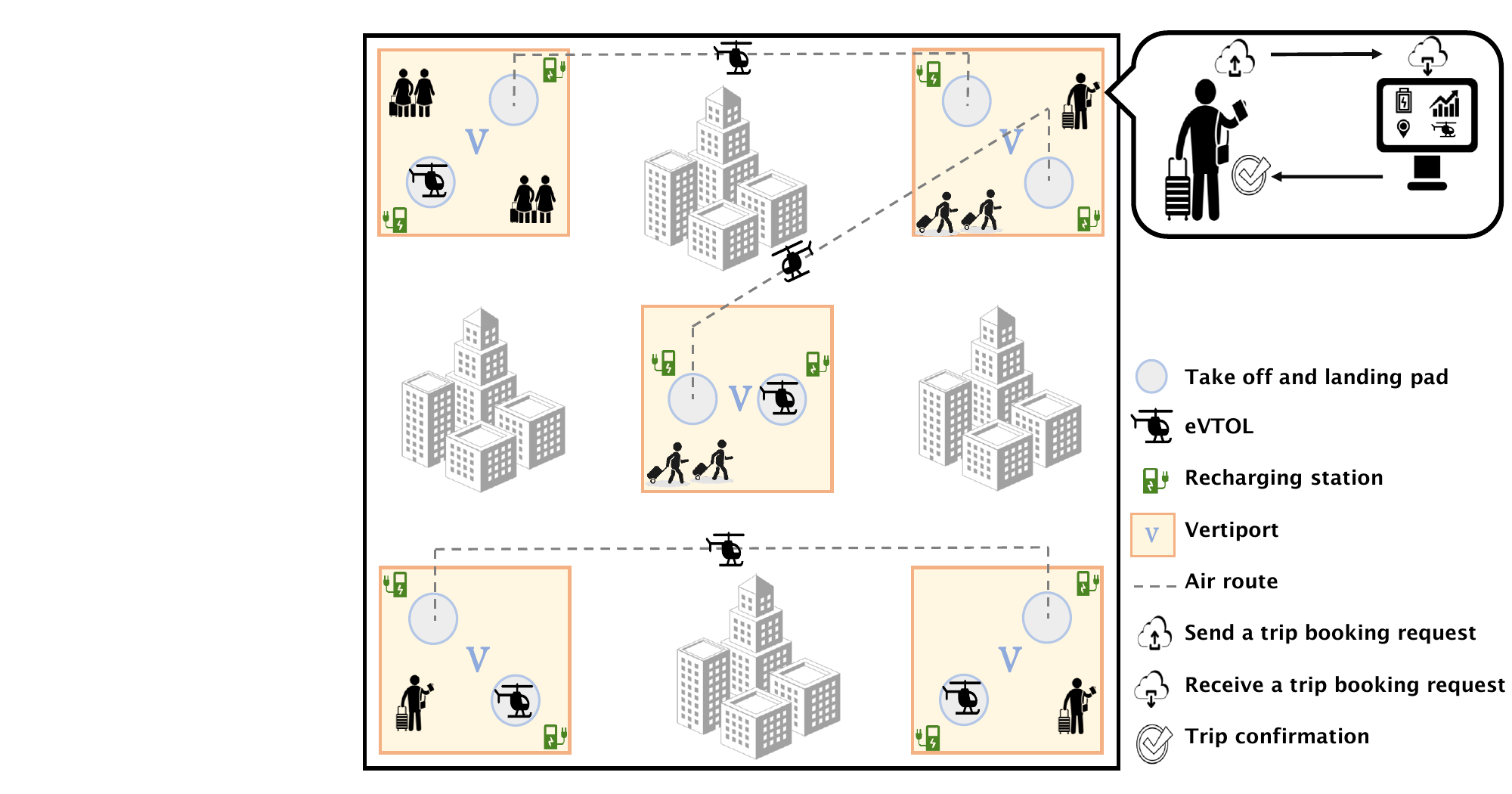}
\caption{The future of commuting: AAM passenger transportation}
\label{AAM-Figure}
\end{figure}

Figure \ref{AAM-Figure} demonstrates the concept of passenger transportation in AAM. From the passenger's point of view, they would first request rides to the AAM air taxi operator. After receiving trip confirmation, they would arrive at the origin vertiport before their flight time, and fly to their destination vertiport via an eVTOL at the flight departure time. On the other side, from the AAM air taxi operator's perspective, they would first receive trip requests from passengers, then determine how many aircraft and which aircraft to dispatch for transporting groups of ridesharing passengers and to dispatch for recharging, confirm trip requests to passengers accordingly, and then fly the aircraft from the origin vertiport to the destination vertiport at the flight times. 

 \subsection{Related Work and Research Motivation}
 \label{Related Work and Research Motivation}

Since the advent of AAM, it has attracted significant attention from researchers along different directions. These include AAM demand estimation \cite{Demand1, Demand2, vale2023modelling, preis2023estimating}; aircraft design \cite{aircraft1, aircraft2, zhang2023multi}; route planning \cite{path1, path2}; trajectory optimization \cite{traject1, traject2, xiang2022multi, deniz2023autonomous, wu2023precision, wang2023optimal} vertiport siting \cite{vertiport1, vertiport2}; battery technology and charging infrastructure \cite{yang2021challenges}, and communication systems \cite{zaid2021evtol}. For a more comprehensive review of AAM research, interested readers are referred to \cite{review1, review2, review3}. One area that  however still needs to be matured further is aircraft dispatch for AAM passenger transportation. Effective and efficient aircraft dispatch is critical to ensuring profitability and passenger demand fulfillment in AAM passenger transportation operations. Without this, AAM passenger transportation would not be financially sustainable. Therefore, the development of a dispatch algorithm for eVTOLs is necessary with the goal of maximizing operating profit. However, the dispatching of eVTOLs faces several significant challenges which include 1) the need for frequent recharging, 2) limited number of takeoff and landing pads at vertiports for eVTOLs, and 3) time-varying nature of the demand and electricity price. The first two challenges are unique to AAM and not present in other passenger transportation modes. To optimize operating profit, the algorithm needs to dispatch eVTOLs for transporting passengers and recharging based on demand and electricity prices. 


The AAM aircraft dispatch problem is a variant of the general ground vehicle dispatch problem, which arises in numerous applications and involves determining how to optimally dispatch ground vehicles in a given fleet to transport passengers and goods. The optimization objective of the problem varies from minimizing travel time \cite{mint1, mint2}, waiting time \cite{liang2019efficient, belanger2019recent}, energy consumption \cite{energy}, and operating cost \cite{zou2020effective, ocost1} to maximizing the operating profit \cite{maxp1, maxp2, maxp3}, service quality and operation efficiency \cite{serviclevel1, serviclevel2}. Various constraints --- such as maximum ride time, vehicle fuel and passenger capacity, vehicle payload or weight limits, limited request time, and more --- are taken into account depending on the chosen objective function. A comprehensive survey of existing literature on vehicle dispatch problems reveals four different solution approaches \cite{cai2023survey}: optimization \cite{opt1, opt2, belanger2019recent, yin2022optimal, mao2020dispatch}, heuristics \cite{heuristic1, heuristic2}, meta-heuristic \cite{metaheuristic1, metaheuristic2, metaheuristic3, metaheuristic4}, and reinforcement learning (RL)\cite{guo2020deep, holler2019deep, liu2022deep, shi2019operating, feng2021scalable, oda2018movi}.

Compared to the problem of dispatching ground vehicles or taxis to transport passengers, dispatching aerial vehicles like eVTOLs for passenger transportation faces two unique challenges. First, the number of takeoff and landing pads in each vertiport is limited in AAM. Second, eVTOLs have limited battery capacity and hence need frequent recharging. Other challenges which are common to both eVTOL dispatch and electric ground vehicle dispatch include the dynamic demand and electricity prices. In solving the problem of dispatching eVTOLs, two different solution approaches can be found in the literature: optimization and RL. In \cite{shihab2019schedule}, optimization models are proposed to help commercial AAM air taxi operators compare three types of possible dispatching --- namely, on-demand operations, scheduled operations, and hybrid operations --- and determine the dispatch solution based on simulated market demand, such that the operating profit is maximized. Dynamic electricity price was not considered and, hence, electricity price was kept fixed for the whole operating horizon. Another optimization model is presented in \cite{shihab2020optimal} for dispatching eVTOLs. For both passenger transportation and providing power grid services, either independently or together. The authors expanded upon the dispatch model presented in \cite{shihab2019schedule}. This expansion includes incorporating a more realistic charging process with time-varying electricity prices, a battery model with degradation, and the option of providing frequency regulation service to the power grid to gain additional revenue. It investigated the trade-off among multiple revenue and cost sources for eVTOL fleet operations, with the following multiple objectives: 1) maximizing the revenue generated from transporting passengers, 2) maximizing the revenue generated from providing frequency regulation services to the power grid, and 3) reducing the operating and charging costs. In \cite{roy2022flight}, a multi-commodity network flow framework is developed to optimize flight scheduling for an airport shuttle air taxi service, assessing its profitability based on factors like population density, fleet size, ticket price, and operating costs. Although the optimization approach can determine the exact maximum operating profit over a given operating horizon, modeling the objective function and the operational constraints can be challenging. Also, the run times of the optimization models becomes prohibitively large with the number of vertiports and eVTOLs and the operating horizon considered in the problem. This is particularly challenging for on-demand operations, where the models need to follow a rolling horizon approach and need to be rerun every time the parameters in the operating environment (e.g., demand, electricity price) change \cite{shihab2020optimal, shihab2019schedule}. 

Approaches based on heuristics and meta-heuristics typically have much lower computational run times than their optimization counterparts, however, they too typically require formulation of the optimization model \cite{cai2023survey}. The RL approach can potentially overcome both the limitations of the optimization approach as 1) they are inherently model-free and 2) the training run time of RL approach during deployment is significantly smaller than that of the optimization approach. In theory, the RL approach can converge to optimal solutions given sufficient training time and, in practice, it typically yields near-optimal solutions with very small optimality gaps \cite{sutton2018reinforcement}. So far, \cite{paul2022graph} is the only work which used DRL-based algorithm for AAM fleet scheduling. However, it did not consider several critical and practical AAM operational factors. These include: 1) recharging of eVTOLs, 2) time-varying electricity prices, and 3) limited eVTOL takeoff and landing pads in each vertiport. Without considering these challenges, the AAM dispatch problem becomes incomplete and impractical. Furthermore, no benchmark algorithms or models (e.g., optimization, heuristics, meta-heuristics)  were implemented to compare the performance of their DRL-based algorithm with other approaches.

\subsection{Contributions}

To address these gaps in the literature, the present work employs two different DRL-based approaches --- namely, a single-agent and a multi-agent DRL --- for tackling the eVTOL dispatch problem for AAM passenger transportation. Operational factors and constraints unique to this problem, overlooked in the literature, are taken into account. Real-world commuter data have been used to estimate passenger demand. The performance of both DRL-based algorithms have been benchmarked against an AAM dispatch optimization model to determine how closely the achieved operating profit aligns with the optimal solution. Our contributions are summarized below:

\begin{enumerate}

    \item A single-agent DRL-based algorithm and a multi-agent DRL-based algorithm are developed to solve the problem of dispatching a fleet of eVTOLs with the goal of maximizing the operating profit. 
    
    \item A simulation environment for AAM passenger transportation via eVTOL dispatching is developed considering the relevant operational aspects of the eVTOL dispatch problem, such as limited number of takeoff and landing pads at vertiports, limited eVTOL battery capacities, and time-varying passenger demand and electricity prices.

    \item A series of numerical experiments have been conducted to evaluate the optimality, scalability, and robustness of the algorithms under different demand and electricity price conditions, number of eVTOLs in the fleet, and number of vertiports in the network. Real world electricity prices and estimated AAM passenger demand corresponding to the region selected have been considered in the experiments.  The optimization model given in \cite{shihab2020optimal} have been implemented to benchmark the performance of the DRL-based eVTOL dispatch algorithms to evaluate how closely these DRL-based algorithms could approximate the optimal solution.

\end{enumerate}

The rest of the paper is structured as follows. In Section \ref{Problem Statement}, the overall framework and the formulation of the eVTOL dispatch problem are presented. Section \ref{Solution Methods} elaborates on the single-agent and multi-agent DRL algorithms that we have developed. To validate the effectiveness of our proposed algorithms, Section \ref{Numerical Experiments and Results} presents the simulated environment and a comparative analysis of all three algorithms is in \ref{result}. Finally, the paper concludes with a summary of the research findings and next steps in \ref{Conclusion}.

\section{eVTOL Dispatch Problem Statement}
\label{Problem Statement}

The goal of the eVTOL dispatch problem is to maximize the operating profit of an eVTOL operator over a dispatch horizon $T$ by optimally dispatching a fleet of $N$ eVTOLs across a network of $M$ vertiports to transport passengers. Each episode consists of $n$ time steps of duration $\tau$. In each time step, the eVTOLs are considered to be able to complete their flights along any route within the set of feasible routes between vertiports, denoted by $R_{f}$, and recharge completely. The operating profit is equal to the transportation revenue minus the eVTOL recharging cost and the operating cost. When passengers are transported by an eVTOL from one vertiport to another, they are charged a trip fare per mile of $\rho$, which leads to generation of revenue for the eVTOL operator. The revenue generated from transporting passengers on a route is determined by multiplying three factors: the total number of passengers transported, the length of that route, and the trip fare. During an eVTOL flight, the eVTOL operator incurs an operating cost per mile of $C_{o}$, which is calculated by multiplying the number of available seats by the length of the route. When eVTOLs are recharged at a vertiport charging station, the operator incurs a recharging cost which depends on ${E(t)}$, the electricity price at time step $t$, and it is specified by multiplying electricity price by the amount of battery it requires to be fully charged.

One of three dispatch actions can be taken for each eVTOL during each time step: 1) dispatch it along a feasible route to transport passengers or reposition it (transport action); 2) recharge its batteries at current vertiport location (recharge action); 3) wait at current vertiport location (wait action). For the transport action, the number of passengers transported via eVTOLs depend on passenger demand and passenger capacity of eVTOLs. Let the total passenger demand for each route at time step $t$ be represented by vector $D(t)$. The number of passengers transported by an eVTOL is then bounded by the passenger capacity $C_{p}$ of the eVTOL as well as the demand. To take a feasible recharge action, each eVTOL's battery level and the eVTOL battery capacity need to be known. The battery level for an eVTOL $i$ at time step $t$ is denoted by $B_{i}(t)$, which is always less than or equal to maximum battery capacity $B_{max}$ of eVTOLs. Each eVTOL can only be dispatched on a route $r$ if it has enough battery to complete the flight. Each vertiport has a limited number of takeoff and landing pads $v$. Therefore, each eVTOl can transport passengers to a destination if the destination vertiport has an empty takeoff and landing pad.

\section{Solution Methods}
\label{Solution Methods}

To solve the eVTOL dispatch problem, this paper develops two DRL-based algorithms, namely single-agent and multi-agent DRL algorithms. They are described in Sections \ref{Centralized Learning Algorithm}, and \ref{Decentralized Learning Algorithm}, respectively. As discussed in detail in Section \ref{Related Work and Research Motivation}, the motivation behind employing a DRL approach to the eVTOL dispatch problem is to overcome 1) some of the limitations of the optimization-based approaches and 2) current gaps in the literature. 

Deep reinforcement learning combines reinforcement learning with deep neural networks to determine the optimal policy. It formulates the problem as a Markov Decision Process (MDP), with defining state space, action space, and reward functions for the agent(s). The agent(s) gets rewarded according to the action it takes. If it leads to a better outcome, the reward is positive. Otherwise, the reward function discourage the agent with a negative or low reward to repeat that action. The objective of the DRL agent is to find the optimal policy over an episode. Optimal policy is a set of sequential decisions for the agent(s) which lead to attaining the maximum expected cumulative reward. In each time step, the expected cumulative reward for each action is estimated by the deep neural network according to the state of current time step. The following section will provide a comprehensive explanation of the single-agent DRL-based algorithm.

\subsection{Single-agent Deep Reinforcement Learning}
\label{Centralized Learning Algorithm}

In this approach, the fleet of eVTOLs is controlled by a single DRL agent which is aware of the state of all the eVTOLs. It takes actions that define the dispatch decisions for each eVTOL in the fleet. Its objective is to maximize the total cumulative reward obtained by taking actions to achieve the highest possible operating profit over the dispatch horizon.

\subsubsection{Markov Decision Process Formulation}
\label{MDP}
 We model our problem as an MDP with a single agent that acts as the central decision maker. The MDP is denoted by $<S, A, s(0), P, R, \gamma>$, where $S$ is the state space comprising all possible states $s(t)$ in the environment; $s(t)$ is the state of environment at time step $t$; $A$ is the action space comprising all possible actions $a(t)$  available to the agent; $a(t)$ is the action taken by the agent at time step $t$, $s(0)$ is the initial state, $P(s(t+1)|s(t), a(t))$ is the transition function, $R(s(t), a(t))$ is the reward function obtained from state $s(t)$ by taking action $a(t)$, and $\gamma$ is the discount factor. At each time step $t$, the central agent observes a state $s(t) \in S$ and takes an action $a(t) \in A$. The environment then shifts the agent to the next state $s(t+1) \in S$ based on $P(s(t+1)|s(t), a(t))$, and returns the agent the reward $r(t)$ based on $R(s(t), a(t))$. As the dispatch horizon is short for this problem, the discounting of rewards is not particularly relevant for this problem, and, hence, $\gamma$ can be kept one or close to one for this problem. The definitions of the state space, action space, transition function, and reward function of the eVTOL dispatch MDP are discussed below:

\textbf{State Space:} 

The state space is defined to represent the environment of the eVTOL dispatch problem. A four-tuple is formulated to represent the state of the environment at a given time: $s(t) = [t, \hat{s}_{e}(t), E(t), D(t)]$.  $\hat{s}_{e}(t)$ is a tuple representing the state of all eVTOLs: $\hat{s}_{e}(t) = [\hat{s}_{1}(t),...,\hat{s}_{N}(t)]$, where $N$ represents the total number of eVTOLs. $\hat{s}_{i}(t)$ denotes the state vector of the $i$th eVTOL at time step $t$: $\hat{s}_{i}(t) = [B_{i}(t), L_{i}(t)]$, where $B_{i}(t)$ and $L_{i}(t)$ are the battery level and location of the $i$th eVTOL at time step $t$, respectively. $E(t)$ $(\$/KWh)$ denotes the electricity price at time step $t$, and $D(t)$ is a vector that indicates the passenger demand for each feasible route: $D(t) =[d_{12},d_{13}, ..., d_{M (M-1)}]$. Here, $d_{jk}$ represents the passenger demand at time step $t$ for the route starting from the vertiport $j$ and ending at the vertiport $k$ and $M$ the total number of vertiports.

In a deterministic scenario, the range of possible values for each state space variable is finite. Considering a fully connected $M$-vertiport network with $N$ eVTOLs, the potential values for these state variables can be defined. Time step $t$ can adopt any of $n$ discrete values, as one episode consists of $n$ time steps. Similarly, the electricity price $E(t)$ can have $n$ potential values, each corresponding to one of the $n$ time steps in an episode. The passenger demand vector $D(t)$ is a tuple where each element can take on one of $n$ possible values and each value is related to a time step. The battery level $B_{i}(t)$ is a continuous variable and its possible values depend on factors such as the maximum battery capacity, charging rate, discharging rate, and the specific set of feasible routes. The location $L_{i}(t)$ state variable can assume one of $M$ distinct positions within the vertiport network.

\textbf{Action Space:} 

At each time step, for each eVTOL, the agent can take up to $M+1$ actions, with one action being for waiting at $L_{i}(t)$, $M-1$ actions for transporting passengers to vertiports other than $L_{i}(t)$, and one action for recharging at $L_{i}(t)$. Let $\hat{a}_{i}(t)$ denote the action taken by the agent for eVTOL $i$. Then, the actions are defined as below:

\begin{itemize}

  \item Recharge action ($\hat{a}_{i}(t) = 0$): If the agent selects $0$ for eVTOL $i$ ($\hat{a}_{i}(t) = 0$), it means it decides to recharge the battery of eVTOL $i$ at time step $t$. During recharging, the battery level will increase based on the charging rate and time step duration. As discussed previously, we consider the time step duration to be sufficiently long such that the battery can get fully recharged within the time step from any battery level $B_i(t)$. However, if the eVTOL is already fully charged, then this action does not lead to any further increase in the battery level.

  \item Wait action ($\hat{a}_{i}(t) = L_i(t)$): If the agent selects the same vertiport as the current vertiport location for eVTOL $i$ to be dispatched to ($\hat{a}_{i}(t) = L_i(t)$), then eVTOL $i$ waits at $L_i(t)$ for time step $t$.

  \item Transport action ($\hat{a}_{i}(t) = V \setminus L_i(t)$): If the agent selects a destination vertiport other than the current vertiport location for eVTOL $i$ to be dispatched to ($\hat{a}_{i}(t) = V \setminus L_i(t)$), then eVTOL $i$ flies and transports passengers from its origin vertiport $L_i(t)$ to destination vertiport $\hat{a}_{i}(t)$ during time step $t$ if it has sufficient battery capacity for the trip. If the battery level is insufficient for the chosen route, eVTOL $i$ must remain at vertiport $L{i}(t)$ for the duration of time step $t$. For instance, if $L_i(t) = 1$, then $\hat{a}_{i}(t) = 3$ indicates that the eVTOL $i$ is being dispatched from its origin vertiport $1$ to destination vertiport $3$ to transport a given number of passengers up to its capacity. 
  
\end{itemize}

\noindent Hence, the action space $A(t) = [a_{1}(t), ..., a_{{(M+1)}^{N}}(t)]$ and the size of action space (total number of actions in the action space) is ${(M+1)}^{N}$, which increases exponentially with $M$ and $N$. As an example, the action space for the case of two eVTOLs $(N = 2)$ and two vertiports $(M = 2)$ comprises 9 actions, which are listed in Table \ref{table1}. 

\begin{table}[h]
\caption{Action space for the single agent ($M = 2, N = 2$)}
\label{table1}
\centering
\begin{tabular}{@{}lll@{}}
\toprule
\textbf{Action} & \textbf{$\hat{a}_{1}$} & \textbf{$\hat{a}_{2}$} \\ \midrule
$a_{1}$  & \(0\)  & \(0\)\\
$a_{2}$  & \(0\)  & \(1\)\\
$a_{3}$  & \(0\)  & \(2\)\\
$a_{4}$  & \(1\)  & \(0\)\\
$a_{5}$  & \(1\)  & \(1\)\\
$a_{6}$  & \(1\)  & \(2\)\\
$a_{7}$  & \(2\)  & \(0\)\\
$a_{8}$  & \(2\)  & \(1\)\\
$a_{9}$  & \(2\)  & \(2\)\\\bottomrule
\end{tabular}
\end{table}

\newpage
\textbf{Transition Function:}
\label{ST}

At each time step $t$, the agent's state $s(t)$ undergoes updates based on the action $a_{i}(t)$ it takes. For all actions, as the agent transitions from the current state to the next state ($s(t)\rightarrow{s(t+1)}$), time step gets incremented by one ($t\rightarrow{t+1}$). The electricity price and passenger demand get updated to their respective values associated with the next time step. The update of other state tuple elements depend on the specific action taken by the agent:

\begin{itemize}

  \item Recharge action ($\hat{a}_{i}(t) = 0$): If the agent selects a recharge action for eVTOL $i$ at time step $t$, considering that the time step duration is sufficient to fully recharge an empty eVTOL, the battery level of that eVTOL will be set to maximum battery capacity $B_i(t+1) = B_{max}$. eVTOL $i$'s vertiport location remains the same ($L_{i}(t+1) = L_{i}(t)$).

  \item Wait action ($\hat{a}_{i}(t) = L_i(t)$): If the agent selects a wait action for eVTOL $i$ at time step $t$, the state of eVTOL $\hat{s}_{i}(t)$, including its location $L_{i}(t)$ and battery level $B_{i}(t)$ remains unchanged.

  \item  Transport action ($\hat{a}_{i}(t) \in V \setminus L_i(t)$): If the agent selects a transport action for eVTOL $i$ at time step $t$, and the eVTOL transports $w_{jk}(t)$ passengers from vertiport $j$ to vertiport $k$ at time step $t$, the location of the eVTOL at the next time step $t+1$ will be the destination vertiport $k$ ($L_i(t+1) = k$). The battery level of the eVTOL will be updated according to Eq.  \ref{battery}:

\begin{equation}
\label{battery}
B_{i}(t+1) = B_{i}(t) -  \eta  \Tilde{t}_{jk},
\end{equation}. 

Where $\Tilde{t}_{jk}$ is the time takes for an eVTOL to fly from vertiport $j$ to vertiport $k$. If the number of passengers on the route starting from vertiport $j$ and ending at $k$ at time step $t$ is greater than or equal to the passenger capacity of the eVTOL $C$, then $w_{jk}(t)$ is set to the maximum passenger capacity $w_{jk}(t) = C$. Otherwise, $w_{jk}(t)$ is set to the number of passengers on that route at time step $t$. If for eVTOL $i$ and $i+1$ the same transport actions $\hat{a}_{i}(t) = \hat{a}_{i+1}(t) = k$ in the same origin vertiport $L_{i}(t) = L_{i+1}(t) = J$ are taken by the agent, the eVTOL $i$ which is in row is in priority. According to $w_{jk}(t)$ for eVTOL $i$, the number of passengers on that route $r_{jk}$ is updated $d_{jk}(t)$ $\leftarrow$ $d_{jk}(t) - w_{jk}(t)$ for eVTOL $i+1$. 
\end{itemize}

\textbf{Reward:} 

At each time step $t$, the reward $r(t)$ obtained by the agent is the accumulation of the rewards $\hat{r}_{i}(t)$ associated with the actions for each eVTOL: 

\begin{equation}
\label{equation6}
r(t)= \sum_{i=1}^{N} \hat{r}_{i}(t)
\end{equation}

\noindent The total reward achieved by the agent $R$ in an episode of $n$ time steps is calculated as: 

\begin{equation}
\label{total_reward}
R = \sum_{t=1}^{n} r(t)
\end{equation}

\noindent The rewards $\hat{r}_{i}(t)$ associated with the different actions taken for each eVTOL are specified as below:

\begin{itemize}
  \item Recharge action ($\hat{a}_{i}(t) = 0$): The recharging cost, and, hence, the reward associated with recharge action for a given eVTOL $i$ in a given time step $t$ is determined by the amount of electric energy the eVTOL receives from the power grid $(B_{max} - B_{i}(t))$ during that time step and the electricity price $E(t)$ at that particular time step. The reward corresponding to the recharge action is given by: 
 \begin{equation}
\label{equation7}
\hat{r}_{i}(t)  = - (\frac{(B_{max} - B_{i}(t))* E(t)}{B_{max} * {E_{max}}}),
\end{equation}

where $E_{max}$ is maximum electricity price over the entire episode. Higher electricity prices and larger amount of energy consumption from the grid result in higher recharging costs, and, hence, lower rewards. The reward function for recharge action is normalized by dividing the reward function with ($B_{max} * {E_{max}}$) to limit it within a certain range. 
  
  \item Wait action ($\hat{a}_{i}(t) = L_{i}(t)$): The reward for a wait action for any eVTOL in a time step is zero as neither any revenue is generated nor any costs are incurred from this action.

  \item Transport action ($\hat{a}_{i}(t) \in V \setminus L_{i}(t)$): The profit obtained, and, hence, the reward received from a transport action for eVTOL $i$ in time step $t$  depends on the revenue generated from passenger transportation and flight operating cost. The revenue generated from the flight ($w_{jk}(t)*\rho*l_{jk}$) is proportional to the number of passengers transported $w_{jk}(t)$ from origin vertiport $j$ to destination vertiport $k$, the trip fare $\rho$ dollars per mile per passenger, and the distance $l_{jk}$ from origin vertiport $j$ to destination vertiport $k$. Hence, for a given $\rho$, an eVTOL that is dispatched on a route with a greater distance and more passengers will receive a higher reward. The operating cost ($\rho_{o}*(C+1)*l_{jk}$) depends on operating cost $\rho_{o}$ dollars per mile per available seat, passenger capacity $C$ of eVTOLs, and $l_{jk}$. $C+1$ refers to the number of available seats in an eVTOL with $C$ seats for passengers and a seat for the pilot. Hence, the reward corresponding to transport action for eVTOL $i$ in time step $t$ is specified as:

\begin{equation}
\label{equation10}
\hat{r}_{i}(t) = \frac{((w_{jk}(t)* \rho) - (\rho_{o}*(C+1)))*l_{jk}}{C* l_{max}},
\end{equation}  

where $l_{max}$ is the maximum distance for the routes in $R_{f}$ in the vertiport network. To limit the reward function to a specific range of values, the reward function is normalized using $l_{max}$ and $C$ in its denominator.

\end{itemize}

\subsubsection{Single-agent Deep Q-learning Algorithm}
\label{SDQL}

The Deep Q-Learning (DQL) \cite{mnih2015human} algorithm can be used to solve the single-agent eVTOL dispatch MDP problem given its large state and action spaces. DQL merges the concepts of Deep Learning and Q-learning to tackle the limitations posed by large state and action spaces in conventional Q-learning algorithm. This is achieved in DQL by using a neural network called the Q-network instead of a table to calculate the Q-value of each state-action pair. 

\begin{figure}[h]
\centering
\includegraphics[width=15cm, height = 6.2 cm]{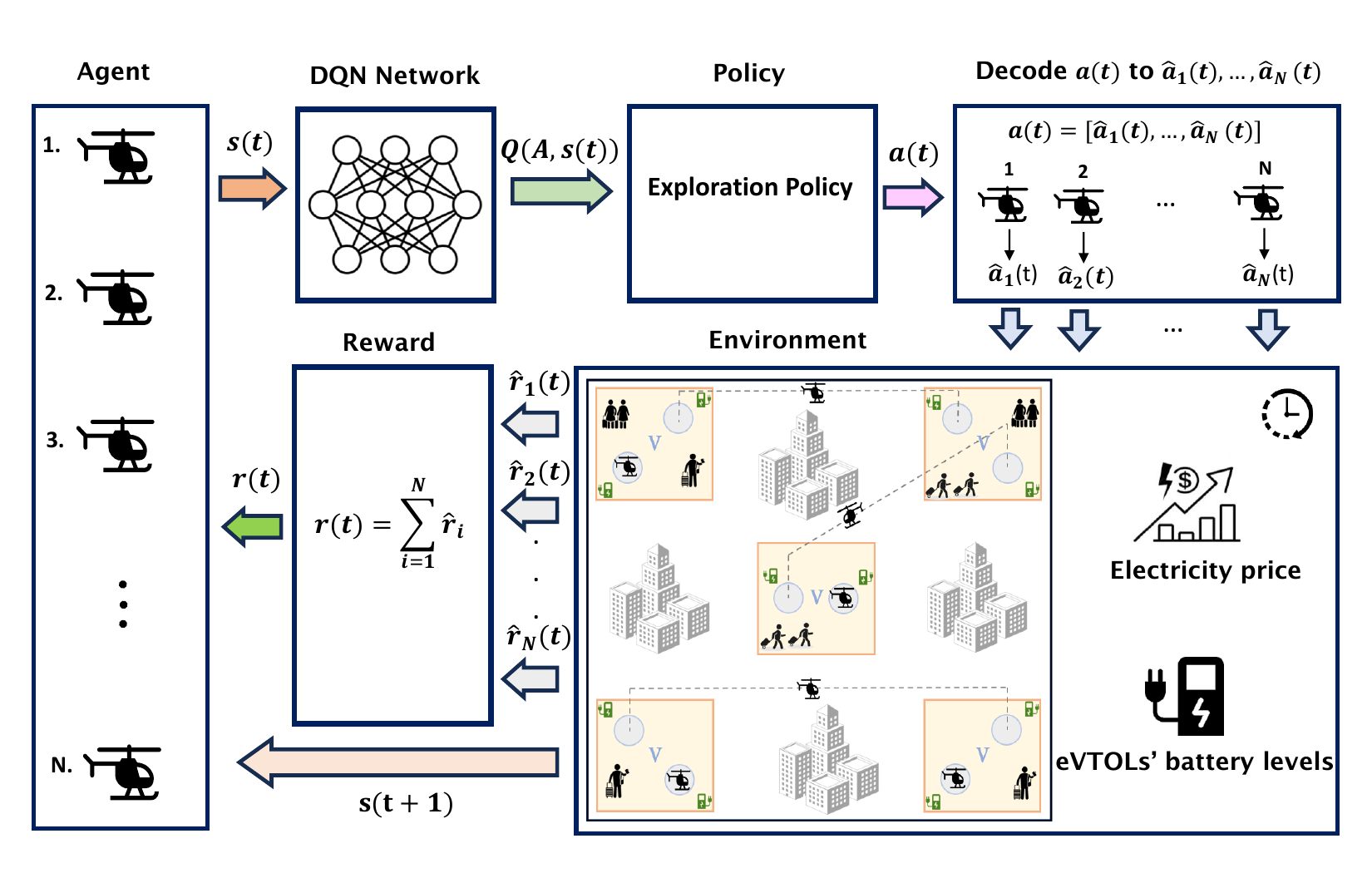}
\caption{Schematic structure of single-agent deep Q-learning eVTOL dispatch algorithm}
\label{Methodology}
\end{figure}

The Q-network algorithm operates by accepting the current state of the agent as its input and producing the Q-values for every action that can be taken. The purpose of the network is to predict the anticipated future rewards for each action taking into account the present state. The Q-network is trained through a process of trial and error where the agent interacts with the environment and adjusts the Q-network based on the rewards observed.
As illustrated in the algorithm's schematic structure given in Fig. \ref{Methodology}, the agent at each time step, selects an action based on the present Q-values by implementing an exploration strategy like epsilon-greedy, which considers exploiting the current optimal action and exploring new actions with probabilities of $1-\epsilon$ and $\epsilon$, respectively. In this context, we utilized the Linear-annealed-policy to make a good trade off between exploiting and exploring. Subsequently, the agent carries out the selected action and perceives the subsequent state and reward. This collected experience is saved in a replay memory and a random assortment or batch of experiences is employed to modify the weights of the Q-network.

The DQL algorithm updates the weights of the Q-network by comparing the target Q-value for each experience with the predicted Q-value, and minimizing the difference between them. The target Q-value is estimated using another Q-network called target Q-network. The weight of target Q-network is updated less frequently compared to the main Q-network called dispatch Q-network. The update process continues until a stopping criterion, such as a maximum number of episodes or desired performance level, is reached. The steps in implementing the single-agent DQL algorithm for dispatching eVTOLs are outlined in Algorithm \ref{CA}.

\begin{algorithm}[H]
        \setstretch{1.0}
	\caption{Single-agent DQL-based eVTOL Dispatch Algorithm}
        \label{CA}
	\begin{algorithmic}[1]
 \State Initialize dispatch Q-network with random weights $\theta$, target dispatch Q-network with random weights $\theta^{\prime}$ $(\theta^{\prime} \leftarrow \theta)$, and replay memory $\mathcal{D}$
 
 \State Set minibatch size $B$, target dispatch Q-network update interval $U$, discount factor $\gamma$, and the number of training episodes $m$

 \State Set exploration policy and initial exploration rate $\epsilon$

		\For {episode $= 1, ..., m$}
		\State Initialize the agent's state $s(0)$
		\For {time step $t = 0, ..., n-1$}
            \State $a(t) \sim \pi(a(t)|s(t))$
		\State Decode the dispatch action $a(t)$ to determine the specific action for each eVTOL ($\hat{a}_{i}(t), i \in \{1, ..., N\}$)
            \State Calculate rewards ($\hat{r}_{i}(t), i \in \{1,..., N\}$) based on $R(s(t), \hat{a})_{i}(t))$
		\State Add up each eVTOL's reward ($\hat{r}_{i}(t), i \in \{1,..., N\}$) to calculate the overall reward $r(t)$ for the agent
            \State Update the next state $s(t+1)$ according to the action taken by the agent ($a(t)$)
            \State Store the transition (state $s(t)$, action $a(t)$, reward $r(t)$, next state $s(t+1)$) in replay memory
		\State Sample a minibatch $B$ of transitions
            \State Utilize the dispatch Q-network to approximate the Q-values associated with the current states
            \State Utilize the target dispatch Q-network to approximate the Q-values of the next states
            \State Compute the target Q-values using the Bellman equation
            \State Update $\theta$ using gradient descent on the loss function
            \If {the current time step is a multiple of $U$}

            \State $\theta^{\prime} \leftarrow \theta$ 

            \EndIf
            \State Set the next state to be the current state $s(t)\leftarrow s(t+1)$

		\EndFor
    \EndFor

	\end{algorithmic} 
\end{algorithm}

\subsection{Multi-agent Deep Q-learning Algorithm}

While the single-agent DQL eVTOL dispatch algorithm can be used to solve the eVTOL dispatch problem, it lacks scalability as the corresponding state and action spaces of the MDP formulation gets computationally large when the number of eVTOLs and vertiports increases. To address this issue, cooperative multi-agent DQL is proposed for the eVTOL dispatch problem. In such a cooperative eVTOL dispatch setting with multi-agent DQL, each eVTOL is an individual agent working together to maximize the team's total reward based on the actions it takes and rewards it receives.

\label{Decentralized Learning Algorithm}
\subsubsection{Multi-agent Markov Decision Process}
  In multi-agent DQL, we model our problem as a multi-agent Markov Decision Process (MMDP). The MMDP is denoted by $<N, S^{\prime}, A^{\prime}, s^{\prime}(0), P^{\prime}, R^{\prime}, \gamma>$, where $N$ represents the number of agents, one for each eVTOL in this problem; $S^{\prime}$ is the state space comprising all possible states in the MMDP environment; $A^{\prime}$ is the action space comprising all possible actions $a^{\prime}(t)$ available to the agents; $s^{\prime}(0)$ is the initial state; $P^{\prime}(s^{\prime}(t+1)|s^{\prime}(t), a^{\prime}(t))$ is the transition function; $R^{\prime}(s^{\prime}(t), a^{\prime}(t))$ is the reward function; and $\gamma$ is the discount factor. As mentioned previously in Section \ref{MDP}, $\gamma$ can be kept one or close to one for this problem. As this is a cooperative MMDP where one AAM air taxi operator owns all the eVTOLs, the environment is fully observable for all agents. At each time step $t$, each agent $i$ observes a state $s^{\prime}_{i}(t) \in S^{\prime}$ and takes an action $a^{\prime}_{i}(t) \in A^{\prime}$. The environment then shifts the agent to the next state $s^{\prime}_{i}(t+1) \in S^{\prime}$ based on $P^{\prime}(s^{\prime}_{i}(t+1)|s^{\prime}_{i}(t), a^{\prime}_{i}(t))$, and returns the agent the reward $r^{\prime}_{i}(t)$ based on $R^{\prime}(s^{\prime}_{i}(t), a^{\prime}_{i}(t))$. The definitions of the state space, action space, transition function, and reward function of the eVTOL dispatch MMDP are discussed below:
  
\textbf{State Space:} 

The state space is defined to represent the environment of the eVTOL dispatch MMDP. Each agent has its own state space $s^{\prime}_{i}(t)$. A five-tuple is formulated to represent the state of the environment for each agent $i$ at a given time $t$: $s^{\prime}_{i}(t) = [t, B(t), L(t), E(t), D(t)]$. Here, $B(t)$ is a vector representing the battery levels of all the agents at time step $t$: $B(t) = [B_{1}(t), ..., B_{N}(t)]$. $L(t)$ is a vector representing the location of all the agents at time step $t$: $L(t) =[L_{1}(t), ..., L_{N}(t)]$. $E(t)$ denotes the electricity price $(\frac{\$}{KWh})$ at time step $t$, and $D(t)$ is the passenger demand vector. 

In a deterministic scenario, the range of possible values for each state space variable is finite. Considering a fully connected $M$-vertiport network with $N$ eVTOLs, the potential values for these state variables can be defined. $t$ can adopt any of $n$ discrete values, as one episode consists of $n$ time steps. Similarly, $E(t)$ can have $n$ potential values, each corresponding to one of the $n$ time steps in an episode. The $B_{i}(t)$, which is an element of tuple $B(t)$, is a continuous variable, and its possible values depend on factors such as the maximum battery capacity, charging rate, discharging rate, and the specific set of feasible routes. The $L_{i}(t)$, which is an element of tuple $L(t)$, can have one of $M$ distinct positions within the vertiport network. $d_{jk}$ which is an element of passenger demand vector, considering maximum passenger demand of $d_{jk}$ is $D_{max}$, can have $(\lceil \frac{D_{max}}{C} \rceil + 1 )*n$ different values. Because in each time step, the passenger demand vector gets updated by each action each eVTOL takes.

\textbf{Action Space:} 

At each time step, each (eVTOL) agent in multi-agent eVTOL dispatch has $M+1$ actions to choose from: one action being for waiting at current vertiport $L_{i}(t)$, $M-1$ actions for transporting passengers to vertiports other than $L_{i}(t)$, and one action for recharging at $L_{i}(t)$. Let $a^{\prime}_{i}(t)$ denote the action taken by the agent $i$ at time step $t$, then, the feasible actions for each agent are defined as below:

\begin{itemize}

  \item Recharge action (${a}^{\prime}_{i}(t) = 0$): If agent $i$ selects action $0$, it means it decides to recharge its battery at present vertiport location $L_{i}(t)$. Over the time step, the battery level is considered to get fully recharged. However, if the agent is already fully charged, then this action does not lead to any further increase in the battery level.
  
  \item Wait action (${a}^{\prime}_{i}(t) = L_{i}(t)$): If agent $i$ selects the same vertiport as the current vertiport location to be dispatched to (${a}^{\prime}_{i}(t) = L_{i}(t)$), then agent $i$ waits at $L_{i}(t)$ for time step $t$.

  \item Transport action (${a}^{\prime}_{i}(t) = V \setminus L_{i}(t)$): If agent $i$ selects a destination vertiport other than the current vertiport location to be dispatched to (${a}^{\prime}_{i}(t) = V \setminus L_{i}(t)$), then agent $i$ makes flight and transports $w_{jk}(t)$ passengers from its origin vertiport $L_{i}(t)$ to destination vertiport $a^{\prime}_{i}(t)$ during time step $t$. $w_{jk}(t)$ is specified based on the approach described in in \ref{ST}.


\end{itemize}

\textbf{Transition Function:}

At each time step $t$, the state $s^{\prime}_{i}(t)$ of agent $i$ updates based on the action $a^{\prime}_{i}(t)$ it takes. For all actions, as the agent transitions from the current state to the next state ($s^{\prime}(t)\rightarrow{s^{\prime}(t+1)}$), time step gets incremented by one ($t\rightarrow{t+1}$). The electricity price and passenger demand get updated to their respective values associated with the next time step. The update of other state tuple elements depend on the specific action taken by the agent:

\begin{itemize}

  \item Recharge action ($a^{\prime}_{i}(t) = 0$): If agent $i$ selects the recharge action at time step $t$, then considering that the time step duration is sufficiently long to fully recharge an empty battery, the battery level of that agent will be set to maximum battery capacity $B_i(t+1) = B_{max}$. Agent $i$'s vertiport location remains the same ($L_{i}(t+1) = L_{i}(t)$).

  \item Wait action ($a^{\prime}_{i}(t) = L_i(t)$): If agent $i$ selects the wait action at time step $t$, its location $L_{i}(t)$ and battery level $B_{i}(t)$ remains unchanged.

  \item Transport action ($a^{\prime}_{i}(t) \in V \setminus L_i(t)$): If agent $i$ selects a transport action at time step $t$ and the agent transports $w_{jk}(t)$ passengers from vertiport $j$ to vertiport $k$, the location of the agent at the next time step $t+1$ will be the destination vertiport $k$ ($L_i(t+1) = k$). The battery level $B_i(t)$ of the agent will be updated according to Eq. \ref{battery}.

\end{itemize}

\textbf{Reward:} 

At each time step $t$, the total reward $r(t)$ obtained by the $N$ agents is the sum of the rewards $r^{\prime}_{i}(t)$ associated with the actions for each agent:

\begin{equation}
\label{multi-reward}
r(t)= \sum_{i=1}^{N} {r}^{\prime}_{i}(t)
\end{equation}

\noindent Then, the total reward $R^{\prime}$ achieved by the agents in an episode of $n$ time steps is calculated as:

\begin{equation}
\label{multi-total-reward}
R^{\prime}(t)= \sum_{t=1}^{n} {r}(t)
\end{equation}

\noindent The rewards $r^{\prime}_{i}(t)$ associated with the different actions taken by agent $i$ at time step $t$ are specified as below:
\begin{itemize}
  \item Recharge action ($a^{\prime}_{i}(t) = 0$): The reward associated with recharge action for agent $i$ $r^{\prime}_{i}(t)$ is determined using Eq. \ref{equation7}.

  \item Wait action ($a^{\prime}_{i}(t) = L_i(t)$): The reward for a wait action for any agent in a time step is zero as neither any revenue is generated nor any costs are created by this action.
  
  \item Transport action ($a^{\prime}_{i}(t) \in V \setminus L_i(t)$): The reward received by agent $i$ $r^{\prime}_{i}(t)$ by taking the transport action is calculated according to Eq. \ref{equation10}

\end{itemize}

\subsubsection{Multi-agent Deep Q-learning Algorithm}

\begin{figure}[h]
\centering
\includegraphics[width=16.5cm, height = 5 cm]{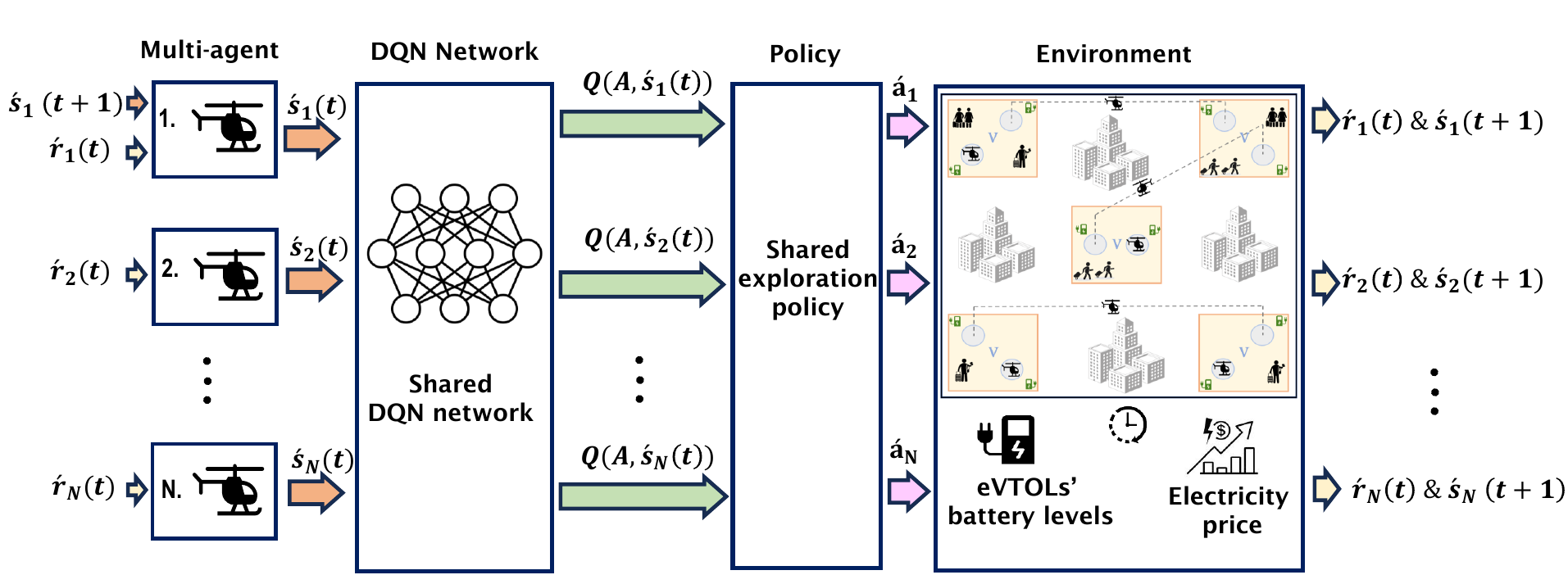}
\caption{Schematic structure of multi-agent deep Q-learning eVTOL dispatch algorithm}
\label{Multi-agent}
\end{figure}

 The multi-agent DQL \cite{mnih2016asynchronous} algorithm can be used to solve the eVTOL dispatch MMDP problem. In this algorithm, each eVTOL acts as an independent DRL agent and it follows the training process described in Section \ref{SDQL}. Agents can either have their own Q-network and own exploration policy or have a shared Q-network and shared exploration policy. We considered the latter setting in the algorithm in this study, as illustrated in the algorithm's schematic structure given in Fig. \ref{Multi-agent}. The use of a shared Q-network and a shared policy reduces the number of model parameters and allows for a more efficient learning process. Compared to the single-agent eVTOL dispatch MDP, the state space of the MMDP is the same as their state tuples are the same;  however, the action space of each agent in the MMDP is significantly smaller as each DRL agent controls only one eVTOL in the MMDP. The steps in implementing the multi-agent DQL algorithm for dispatching eVTOLs are outlined in Algorithm \ref{MA}. The main differences between two DQL eVTOL dispatch algorithms are 1) Instead of having a single agent responsible for all the eVTOLs in single-agent DQL eVTOL dispatch algorithm, each eVTOL is considered to be an individual agent in multi-agent DQL eVTOL dispatch algorithm, and 2) In multi-agent DQL eVTOL dispatch algorithm, agents take actions sequentially, however in single-agent DQL eVTOL dispatch algorithm the agent takes actions for all eVTOLs simultaneously. 


\begin{algorithm}[H]
        \setstretch{1.0}
	\caption{Multi-agent DQL-based eVTOL Dispatch Algorithm}
        \label{MA}
	\begin{algorithmic}[1]
 \State Initialize dispatch Q-network with random weights $\theta$, target dispatch Q-network with random weights $\theta^{\prime}$ $(\theta^{\prime} \leftarrow \theta)$, and replay memory $\mathcal{D}$

 \State Set the number of agents $N$ equal to number of eVTOLs
 \State Set minibatch size $B$, target dispatch Q-network update interval $U$, discount factor $\gamma$, and the number of training episodes $m$ 
 \State Set exploration policy, and the initial exploration rate $\epsilon$

            \For {episode $= 1,..., m$}
		\State Initialize all agents' state $s^{\prime}_{i}(0), i \in \{1,..., N\}$ 
		\For {time step $t = 0,..., n-1$}
            \For{agent $i = 1, ..., N$}
		\State $a^{\prime}_{i}(t) \sim \pi(a^{\prime}_{i}(t)|s^{\prime}_{i}(t))$
            \State Calculate reward $r^{\prime}_{i}(t)$ based on $R^{\prime}(s^{\prime}_{i}(t), a^{\prime}_{i}(t))$
            \State Update all agents' state $s^{\prime}_{i}(t), i \in \{1,..., N\}$ according to $a^{\prime}_{i}(t)$
            \EndFor
            \State Store the transition ($s^{\prime}_{i}(t)$, $a^{\prime}_{i}(t)$, $r^{\prime}_{i}(t)$, $s^{\prime}_{i}(t+1)$) in $\mathcal{D}$
		\State Sample a minibatch of transitions from $\mathcal{D}$
            \State Utilize the dispatch Q-network to approximate the Q-value associated with the current state
            \State Utilize the target dispatch Q-network to approximate the Q-value of the next state
            \State Compute the target Q-values using the Bellman equation
            \State Update $\theta$ using gradient descent on the loss function between target Q-values and Q-values
            \If {the current time step is a multiple of $U$}
            \State $\theta^{\prime} \leftarrow \theta$ 
            \EndIf
            \State $s^{\prime}(t)\leftarrow s^{\prime}(t+1)$ 

		\EndFor
  \EndFor

	\end{algorithmic} 
\end{algorithm}

\newpage
\section{Simulation Environment: Passenger Transportation via eVTOL Dispatching}
\label{Numerical Experiments and Results}
In this section, we present the simulation environment built to train and test the DRL-based eVTOL dispatch algorithms. We start by describing the vertiport network, including the vertiports chosen and the available routes in Section \ref{Vertiport}. Then, we discuss the estimated time-varying AAM passenger demand across the routes in the vertiport network in Section \ref{Passenger}. Section \ref{Aircraft1} introduces the eVTOL model and its specifications used in this study. Then, we provide the time-varying electricity price data specific to the considered vertiport network region in Section \ref{Electricity}.

\subsection{Vertiport Network}
\label{Vertiport}

\begin{figure}[t]
    \centering
    \includegraphics[width=10cm,height=5.5cm]{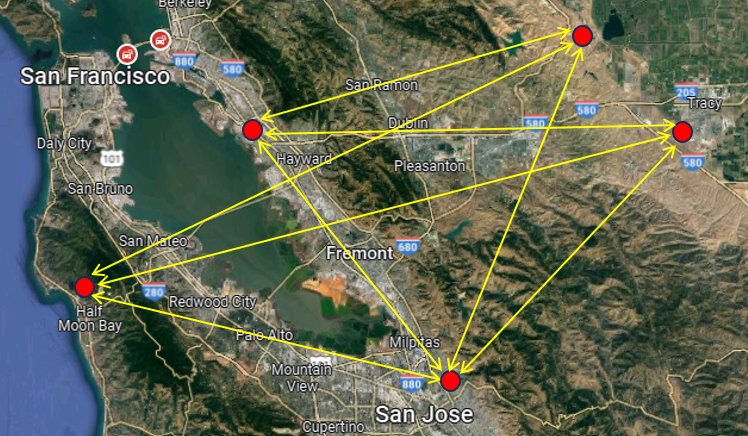}
    \caption{Vertiport network in San Francisco Bay Area in California for AAM passenger transportation}
    \label{fig:vertiport_network}
\end{figure}

In this study, we consider the San Francisco Bay Area in California as the operational region for AAM, given its high potential for AAM passenger transportation. The region consistently ranks among the top five congested areas worldwide \cite{lian2019scalable}. The region is also home to a large number of individuals with sufficiently high income who can afford AAM passenger transportation \cite{rimjha2021commuter}. These factors make it a prime candidate for AAM implementation, given the pressing need for effective solutions to tackle urban traffic congestion challenges. To strategically set up our vertiport network, we choose five counties within this region with the highest number of commuting trips based on the Longitudinal Employer-Household Dynamics Origin-Destination Employment Statistics (LODES) 2019 data available in \cite{LODES}. The counties selected are Santa Clara, San Mateo, Alameda, San Joaquin, and Contra Costa. These counties represent areas where AAM can have a significant impact in potentially alleviating existing congestion issues. Furthermore, these counties are located within a 60-mile radius of any Bay Area county, which is within with the range of the selected eVTOL aircraft as mentioned in Section \ref{aircraft}. We select the locations for vertiports by considering the existing airports within the selected counties, as illustrated in Fig. \ref{fig:vertiport_network} with red circles, while the yellow arrows show the available routes between vertiports. Reid-Hillview of Santa Clara Country Airport, Half Moon Bay Airport, Metropolitan Oakland International Airport, Tracy Municipal Airport, and Byron Airport are five airports we considered as vertiports. Smaller airports with lower air traffic volumes were considered in the vertiport network instead of busier, major airports as the former are more likely to be able to readily accommodate AAM traffic without affecting existing conventional air traffic \cite{preis2021quick}. Among the five vertiports considered in the study, not all of them have routes connecting to every other vertiport, as the demand for eVTOL trips between certain origin-destination vertiport pairs is not significant enough to justify considering routes for those pairs, as discussed below in Section \ref{Passenger}. 

\subsection{Passenger Demand}
\label{Passenger}


We adopt a mixed conditional logit model, as outlined in \cite{rimjha2021commuter}, where the objective was to estimate passenger demand for eVTOL trips specifically in the Bay Area. This model is a discrete choice framework that models how individuals are likely to select their transportation mode based on a utility function associated with each mode, such as car, train, bus, and eVTOL. The utility function is a scalar function that integrates mode-specific attributes like door-to-door trip time and cost, as well as personal characteristics such as income, and potential interaction terms \cite{roy2021user}. Our utility function encompasses four key components: the actual trip cost or fare; the time associated with the trip; the number of required transfers during the trip; and the income level of individuals within a county. The trip fare considered for eVTOL mode is \$4 per mile. This trip time consists of both in-vehicle travel time (IVTT) and out-of-vehicle travel time (OVTT). Here, IVTT denotes the time spent traveling within the chosen mode of transportation, while OVTT refers to the time spent in activities outside the transportation mode, such as waiting for transit or walking between stops. There is a trade-off between the cost of a trip and its duration, which forms the basis for willingness to pay extra for time saved when choosing a faster mode of transportation. Additionally, an individual's mode choice is influenced by the Value of Time (VOT), often assessed with respect to the income level. As mentioned in \cite{borjesson2012income}, higher-income individuals typically have a greater VOT, which leads them to choose faster modes of transportation, prioritizing trip time. Conversely, individuals with lower income levels tend to have a lower VOT and opt for less expensive modes of transportation, prioritizing trip cost. Our model categorizes the selected counties into three income levels: low-income, mid-income, and high-income, based on the data from \cite{rimjha2021commuter}.

Following the approach outlined in \cite{rimjha2021commuter}, we calculate the probability that an individual will choose eVTOL over all available transportation modes within a county. To estimate the number of daily commuting trips from home to work and from work to home for the selected counties, we utilize the LODES-2019 data. We then determine the daily demand for eVTOL trips in a selected county by multiplying the total number of daily commuting trips in that county by the probability that an individual from that county will opt for eVTOL. To accurately represent the actual patterns of commuting trips, we derived the departure time distribution of commuters from National Household Travel Survey data \cite{NHTS}. By utilizing the cumulative density function of departure time distribution of commuters, the demand for eVTOL trips in a specific county is spread throughout the day. Approximately 63\% of these trips are anticipated to take place during the core commuting hours, spanning from 5:30 AM to 8:30 AM and from 4:00 PM to 6:00 PM.

\begin{figure}[t]
\centering
\includegraphics[width=16.5cm, height = 6 cm]{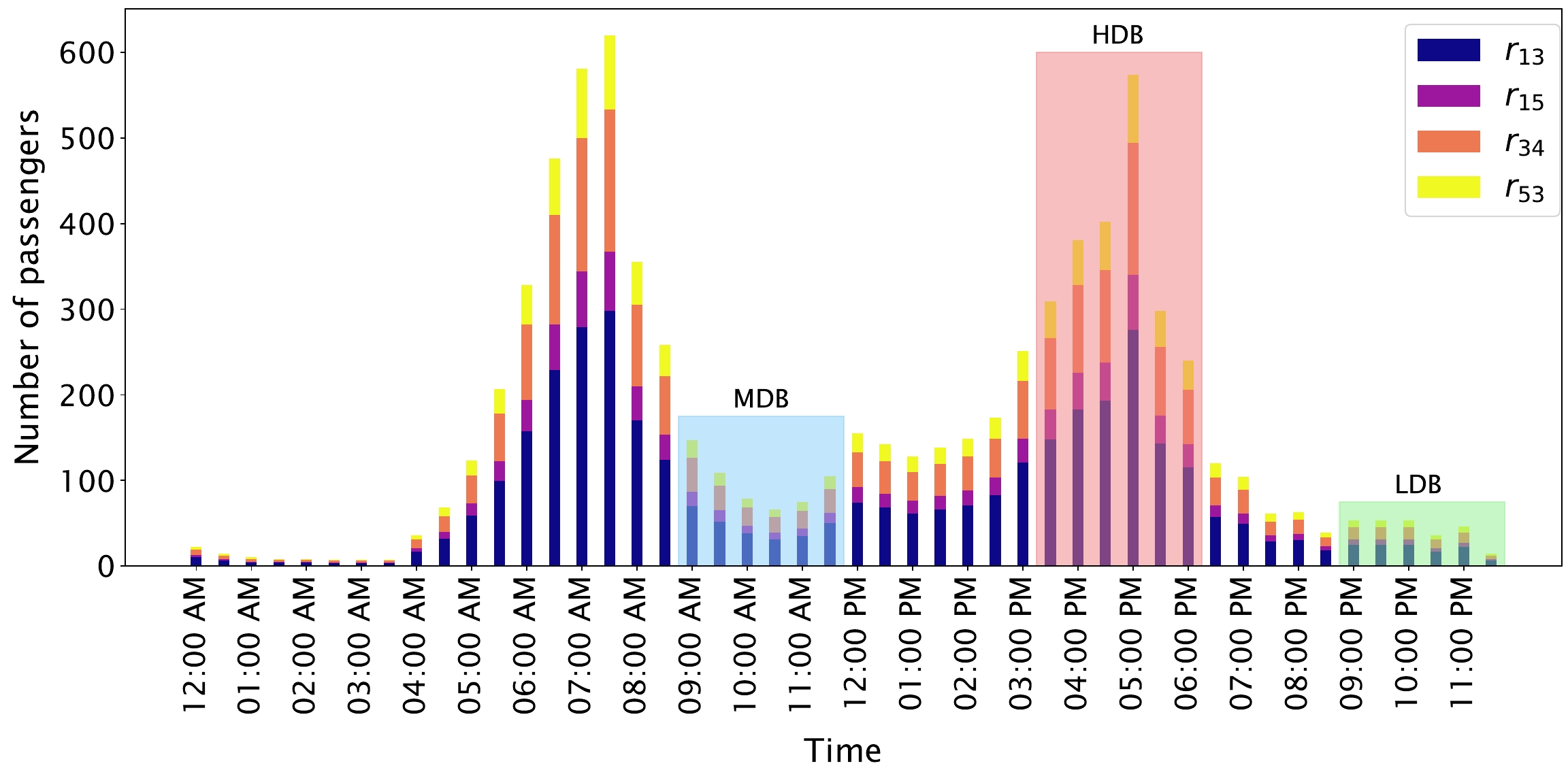}
\caption{Stacked bar chart of potential passenger demand in the various routes at different times of a day}
\label{Total-passenger}
\end{figure}

The vertiports Reid-Hillview of Santa Clara County Airport, Half Moon Bay Airport, Metropolitan Oakland International Airport, Tracy Municipal Airport, and Byron Airport are henceforth referred to as vertiports $1$ through $5$, respectively. A route between vertiports $i$ and $j$ is referred to as $r_{ij}$.  Figure \ref{Total-passenger} illustrates the potential passenger demand for several routes in the selected vertiport network for different times of a given operating day. In a typical day, different levels of passenger demand are anticipated during different times of the day as shown in Fig. \ref{Total-passenger}. To evaluate the performance of our eVTOL dispatch algorithms under different demand settings, we have categorized demand blocks based on varying passenger demand levels. Three demand blocks of interest include the low demand block (LDB), medium demand block (MDB), and high demand block (HDB), as indicated in Fig. \ref{Total-passenger}. The three demand blocks span the time period from 9:00 PM to 11:30 PM, 9:00 AM to 11:30 AM, and 3:30 PM to 6:00 PM, respectively.

\subsection{eVTOL Characteristics} \label{aircraft}
\label{Aircraft1}

\begin{figure}[t] 
    \centering
    \includegraphics[width=10cm,height=4cm]{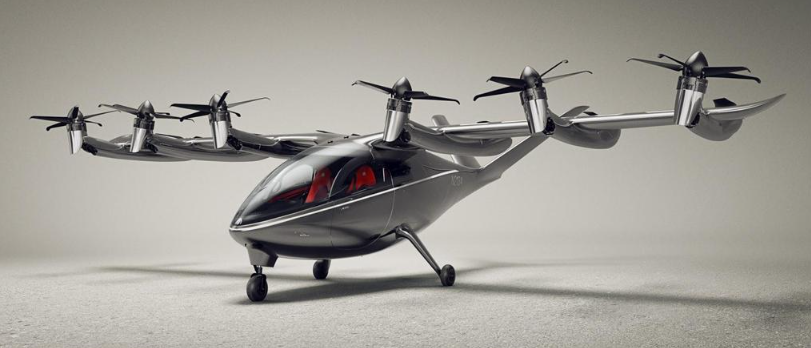}
    \caption{The Archer Aviation eVTOL aircraft}
    \label{vertiport_network}
\end{figure}

The aircraft model used in this study is the eVTOL by Archer Aviation \cite{eVTOLtype}, pictured in Fig. \ref{vertiport_network}. Archer Aviation, headquartered in Palo Alto, achieved a significant milestone by securing a major order worth \$1 billion from United Airlines, making them the first eVTOL company to secure such a substantial order \cite{eVTOL2}. This accomplishment factored into our decision to select this eVTOL for our study. The eVTOL is powered by an electric tilt rotor system, and accommodates one pilot and up to four passengers \cite{eVTOL1}. Following its technical specifications, the range of eVTOL in our study is set to 60 miles, cruise speed to 150 mph, and battery capacity to 140 kWh. The eVTOL operating cost per passenger-mile is considered to be \$1, as reported in \cite{cost}. This estimation is based on various cost factors, including aircraft maintenance, infrastructure, piloting, and other related expenditures.

\subsection{Electricity Price}
\label{Electricity}

\begin{figure}[h]
\centering
\includegraphics[width=16.5cm, height = 6.5 cm]{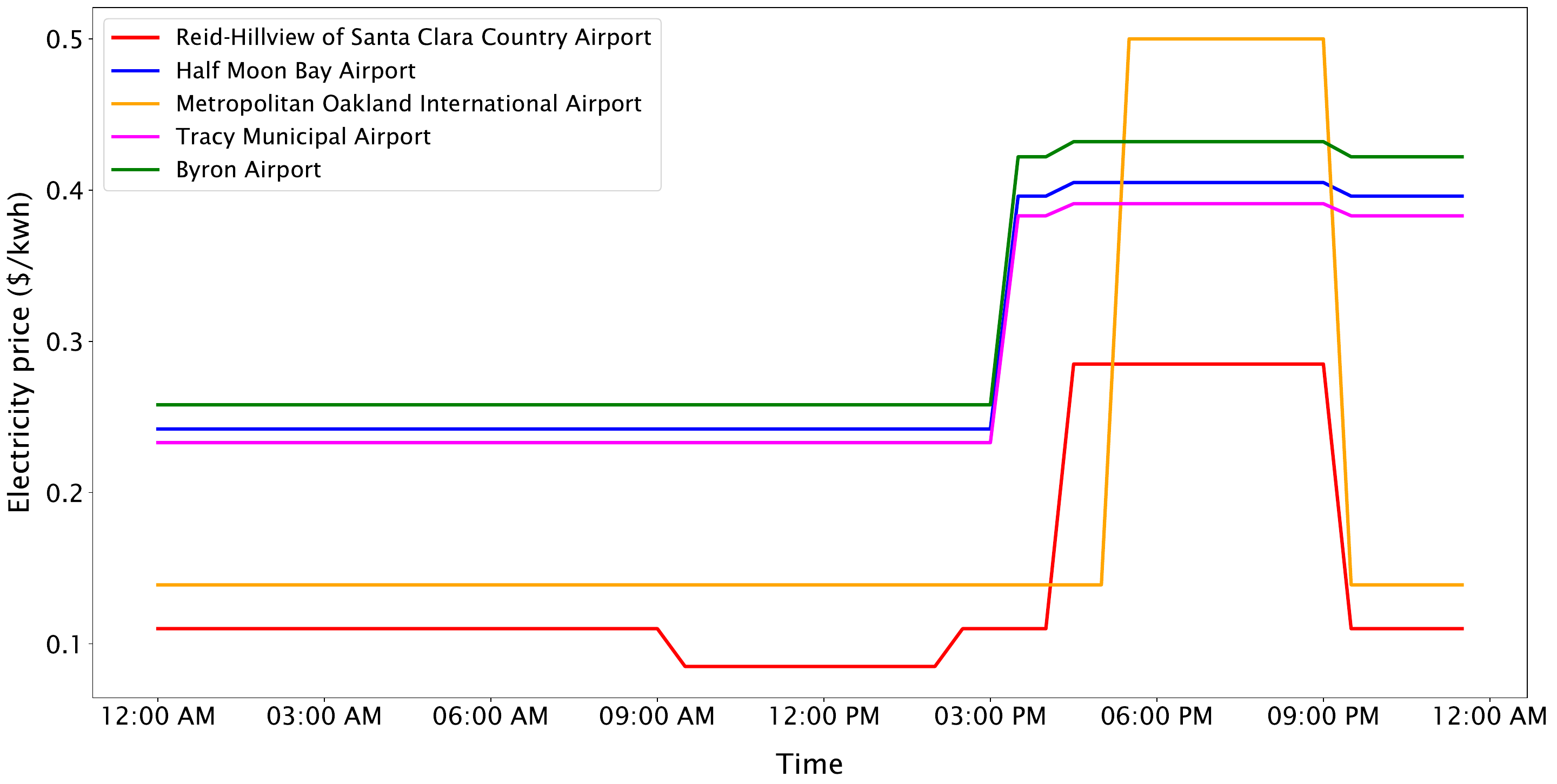}
\caption{Average electricity prices in the Bay Area on June 5, 2023}
\label{EP}
\end{figure}

In the Bay Area, there are different electricity suppliers to provide electricity for that region. During peak power grid load times, their electricity prices tends to be higher, whereas it is comparatively lower during off-peak periods. Although multiple providers exist for the region, their electricity prices at different times of the day tend to align closely. As the different suppliers approximately follow the same pattern and values for electricity pricing, we take the average of their electricity prices and consider it as the electricity price at that time step. We collected the electricity pricing data from the following suppliers in the region: Pacific Gas and Electric Company \cite{PGE}, Alameda Municipal Power \cite{AMP}, and Silicon Valley Clean Energy \cite{SVCE}. The electricity prices at the selected vertiports on June 5, 2023 are shown in Fig. \ref{EP}.

\section{Numerical Experiments and Results}
\label{result}

A series of numerical experiments are conducted in the simulation environment to evaluate the performance of the single agent and multi-agent DQL eVTOL dispatch algorithms. Their performances are benchmarked against the optimal solution obtained by the AAM dispatch optimization model given in \cite{shihab2020optimal}. The setup of the experiments are described in Section \ref{E-C} and the results for the eVTOL dispatch optimization model, the multi-agent DQL eVTOL dispatch algorithm, and the single-agent eVTOL dispatch algorithm are presented in Sections \ref{RO}, \ref{RMA}, and \ref{RSA}, respectively.

\subsection{Experimental Configuration}
\label{E-C}

To evaluate the performance of the algorithms under different demand settings, we considered three different passenger demand blocks as discussed previously in Section \ref{Passenger}: high, medium, and low demand blocks. Each block consists of six time steps ($n = 6$), and the time interval of each time step is $30$ minutes ($\tau = 0.5$). As a result, the time interval for the dispatch horizon is set to $3$ hours ($T = 3$). Each passenger demand block corresponds to a different time of the day, and as a result, the electricity prices considered for each block differ. We varied the number of eVTOLs from $N = 5$ to $N = 20$ in steps of $5$: $N \in \{5, 10, 15, 20\}$. We considered the following values for the number of vertiports: $M = 3$ (vertiports 1, 2, and 3), $M = 4$ (vertiports 1, 2, 3, and 4) and $M = 5$ (vertiports 1, 2, 3, 4, and 5). The number of takeoff and landing pads at each vertiport considered is set to equal to $\lceil N/2 \rceil$. Each combination of $M$ and $N$ was tested under the high, medium, and low demand blocks. In total, there were $36$ numerical cases. For both eVTOL dispatch DQL algorithms and the optimization model, the inital position distribution of eVTOLs were kept the same in all cases. The initial positions were determined using the optimization model. All eVTOLs were set to be fully charged at the beginning of the dispatch horizon. 

\begin{table}[b]
\caption{Hyperparameter settings of DQL algorithms}
\label{tab:configuration}
\centering
\begin{tabular}{@{}lll@{}}
\toprule
\textbf{Hyperparameter}  & \textbf{Value (single-agent)} & \textbf{Value (multi-agent)} \\\midrule
Learning rate & \(0.00001\) & \(0.0005\)\\
Discount factor & \(0.99\) & \(0.99\) \\
Exploration policy & Linear Annealed Policy & Linear Annealed Policy  \\
Initial exploration rate & \(1\) & \(1\) \\
Final exploration rate & \(0.02\) & \(0.02\) \\
Replay buffer size & 50000 & 50000 \\
Minibatch size & 32 & 32 \\
Target Q-network update interval & 500 & 500 \\
Number of hidden layers & 3 [128, 128, 128] & 1 [256] \\\midrule
\end{tabular}
\end{table}

We implemented the single-agent and multi-agent DQL eVTOL dispatch algorithms in Python using Keras-RL \cite{Keras} and the Ray library \cite{Ray}, respectively. Their hyperparameter setting is given in Table \ref{tab:configuration}. Both DQL eVTOL dispatch algorithms were trained and tested on a computer equipped with an 11th Gen Intel(R) Core(TM) i5-1135G7 processor (2.40 GHz) and 20 GiB of memory. The eVTOL dispatch optimization model was implemented in MATLAB and solved using Gurobi 9.1.2 with default settings. As the run times of the optimization model was prohibitively long, especially for experiments involving larger number of eVTOLs and vertiports, the model had to be run on Ohio Super Computer \cite{OSC} with 28 cores.

\subsection{Benchmark Optimization Model for eVTOL Dispatch}
\label{RO}

We implemented the eVTOL dispatch optimization model with a dispatch horizon of six time steps to benchmark the performance of the two DRL-based eVTOL dispatch algorithms. The optimization model was run for all $36$ scenarios, and the maximum operating profits ($\$$) achieved by the model for each numerical case are presented in Table \ref{tab:Profit-optimization}. In each cell of the table, the first, second, and third numbers represent the respective operating profits for the high, medium, and low demand blocks. As seen in Table \ref{tab:Profit-optimization}, both using a larger number of eVTOLs and operating in a larger vertiport network with more vertiports and routes result in higher profits. When the number of eVTOLs increases for a given number of vertiports $M$, more passenger demand can be fulfilled across the routes with a larger fleet of eVTOLs, thereby generating higher operating profit. Also, at any given time step, a larger number of eVTOLs with sufficient levels of charge are available on average for passenger transportation, resulting in higher profits. On the other hand, when the number of vertiports increases for a constant number of eVTOLs $N$, the eVTOLs have a larger number of available routes with passenger demand to choose from for transporting passengers at each time step, leading to higher profits. It can also be observed that the operating profit tends to be higher for higher demand blocks in most cases. These trends concerning the operating profit also hold true for the DQL eVTOL dispatch algorithms, as discussed in the following sections.  


\begin{table}[!h]
\caption{Operating profit (\$) achieved by the eVTOL dispatch optimization model}
\label{tab:Profit-optimization}
\centering
\begin{tabular}{@{}llllll@{}}
\toprule
\textbf{} & & \textbf{$N = 5$} & \textbf{$N = 10$} & \textbf{$N = 15$} & \textbf{$N = 20$} \\\bottomrule
       & HDB & 7407.49  & 14814.99 & 22222.49  & 29630.01\\
$M = 3$  & MDB & 7439.36  & 14878.73 & 22318.08  & 29015.14\\
       & LDB & 7415.31  & 13282.70 & 17002.21  & 18224.98\\\midrule

       & HDB & 9576.23  & 17887.45 & 26208.42  & 34521.17\\
$M = 4$  & MDB & 8363.5  & 16727.01 & 25090.52  & 33034.65\\
       & LDB & 8323.72  & 16092.26 & 23336.76 & 28607.13\\\midrule

       & HDB & 10097.41  & 19737.2 & 28834.45 & 37568.98\\
$M = 5$  & MDB & 9029.38  & 17606.52 & 25961.96  & 34317.45\\
       & LDB & 8463.99  & 16494.97 & 23865.23  & 30514.28\\\bottomrule
\end{tabular}
\end{table}

\subsection{Multi-agent Deep Q-learning eVTOL Dispatch Algorithm}
\label{RMA}

In this section, we evaluate the performance of multi-agent DQL eVTOL dispatch algorithm for the $36$ different numerical cases. The initial positions of the eVTOLs are specified to be the same in both DQL algorithms as that determined by the optimization model. The performance of the algorithm in each case is compared with the result of optimization. The operating profit achieved by the multi-agent DQL eVTOL dispatch algorithm is presented in Table \ref{tab:Profit-DRL}. As observed for the optimization model, the profits achieved by the multi-agent DQL algorithm also increases with the number of eVTOLs and vertiports. Table \ref{tab:error} shows the optimality gap of the multi-agent DQL eVTOL dispatch algorithm with respect to the optimization model in all the numerical experiments. The optimality gaps were observed to be very small, ranging between 0.23 \% and 4.05 \%, and the average optimality gap across all experiments is 1.75 \%. This indicates that the multi-agent DQL eVTOL dispatch algorithm can produce close to optimal solutions. It can also be observed that the optimality gap increases slightly as the scale of the problem instances increase.

\begin{table}[!h]
\caption{Operating profit (\$) achieved by multi-agent eVTOL dispatch algorithm}
\label{tab:Profit-DRL}
\centering
\begin{tabular}{@{}llllll@{}}
\toprule
\textbf{} & & \textbf{$N = 5$} & \textbf{$N = 10$} & \textbf{$N = 15$} & \textbf{$N = 20$} \\\bottomrule
       & HDB & 7349.58  & 14699.15 & 22048.73  & 29349.9\\
$M = 3$  & MDB & 7416.32  & 14832.63 & 22248.94  & 28433.26\\
       & LDB & 7383.4  & 13196.65 & 16677.20  & 18111.47\\\midrule

       & HDB & 9237.95 & 17250.66 & 25699.17  & 33590.93\\
$M = 4$  & MDB & 8344.59  & 16594.56 & 24564.35  & 31907.19\\
       & LDB & 8295.91  & 15640.94 & 22389.68  & 27564.16\\\midrule

       & HDB & 10071.53  & 19287.71 & 27932.87  & 36298.98\\
$M = 5$  & MDB & 8968.67  & 17312.73 & 25575.89  & 33050.89\\
       & LDB & 8434.59  & 16263.09 & 23420.78  & 29286.22\\\bottomrule
\end{tabular}
\end{table}

\begin{table}[!h]
\caption{Optimality gap of the multi-agent DQL eVTOL dispatch algorithm with respect to the optimization model}
\label{tab:error}
\centering
\begin{tabular}{@{}llllll@{}}
\toprule
\textbf{} & & \textbf{$N = 5$} & \textbf{$N = 10$} & \textbf{$N = 15$} & \textbf{$N = 20$} \\\bottomrule
       & HDB & 0.78\%  & 0.78\% & 0.78\%  & 0.95\% \\
$M = 3$  & MDB & 0.31\%  & 0.31\%  & 0.31\%  & 2.01\% \\
       & LDB & 0.43\%  & 0.65\%  & 1.91\%  & 0.62\% \\\midrule

       & HDB & 3.53\%  & 3.56\%  & 1.94\%  & 2.69\% \\
$M = 4$  & MDB & 0.23\%  & 0.79\% & 2.10\%  & 3.41\% \\
       & LDB & 0.33\%  & 2.80\%  & 4.05\%  & 3.64\% \\\midrule
       
       & HDB & 0.26\%  & 2.28\%  & 3.13\%  & 3.38\% \\
$M = 5$  & MDB & 0.67\%  & 1.67\%  & 1.49\%   & 3.70\% \\
       & LDB & 0.35\%  & 1.40\%  & 1.86\%  & 4.03\% \\\bottomrule
\end{tabular}
\end{table}

To have a better view and deeper understanding of the overall performance of the DRL agents in the multi-agent DQL algorithm and the actions taken by them over the dispatch horizon to maximize operating profit, we present the results of one specific numerical case in details among the $36$ different numerical cases. This particular case features four vertiports and five eVTOLs, with the eVTOL dispatch operations taking place over the the high demand block. Note that similar results and patterns were observed for all other cases. 

\begin{figure}[t]
\centering
\includegraphics[width=16.5cm, height = 5.5 cm]{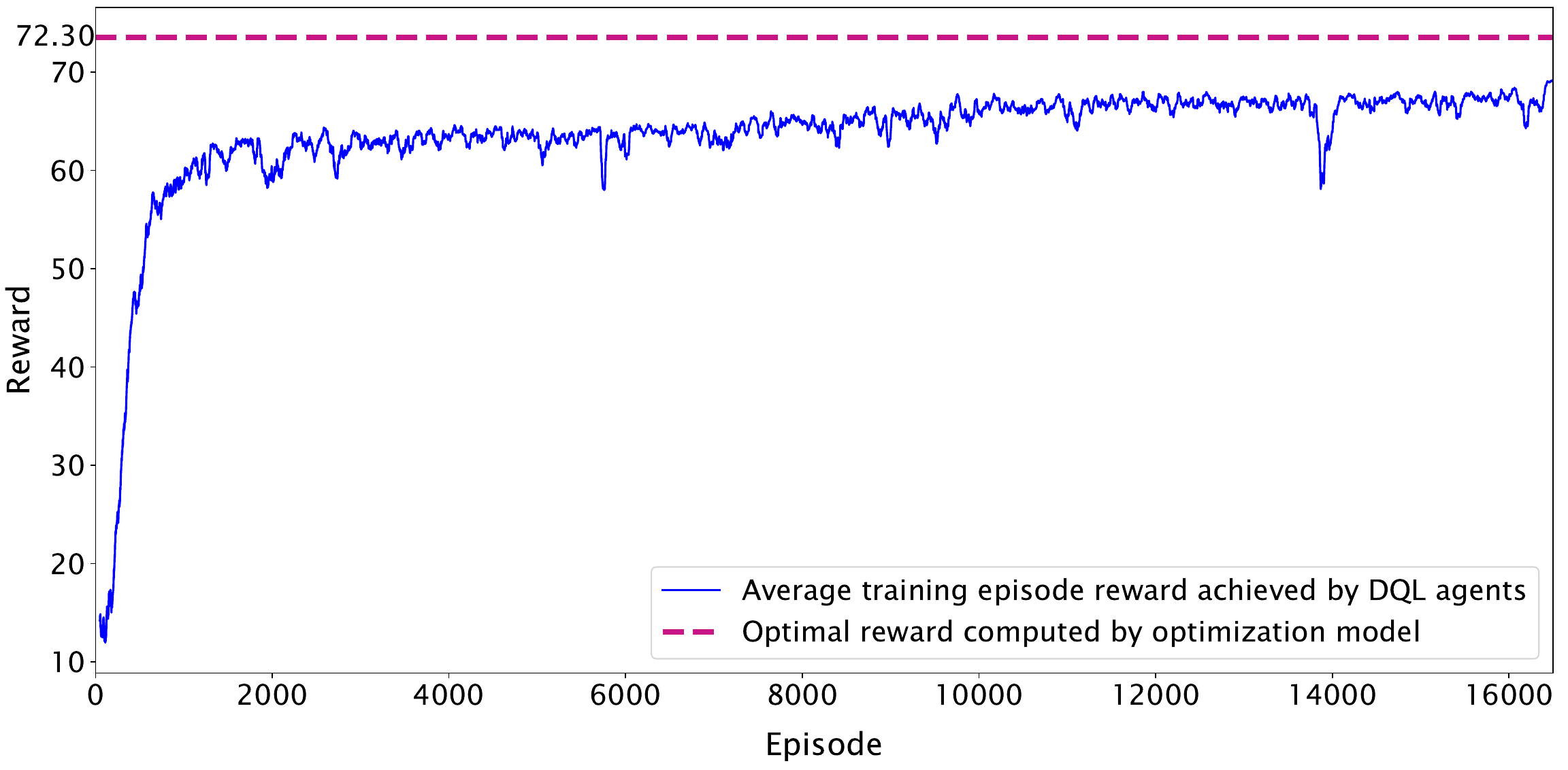}
\caption{Training curve tracking the agents' average episode rewards}
\label{Average-reward}
\end{figure}

To determine the optimal dispatch policy, the agents are first trained in the simulation environment. Fig. \ref{Average-reward} illustrates the training process of the multi-agent DQL eVTOL dispatch algorithm for this specific case. Throughout the training process, the agents explored the state-action space following the linear annealed exploration policy and earned increasing episode rewards. Over the first 1000 episodes, the average collective episodic reward obtained by the agents grew quickly, following which the rewards kept increasing at a slower rate. Through these learning experiences, the agents eventually converged to an optimal dispatch policy at the end of training, achieving a reward of $69.60$, which closely approaches the optimal reward of $72.30$ determined by the eVTOL dispatch optimization model.

\begin{figure}[t]
\centering
\includegraphics[width=16.5 cm, height = 7.5 cm]{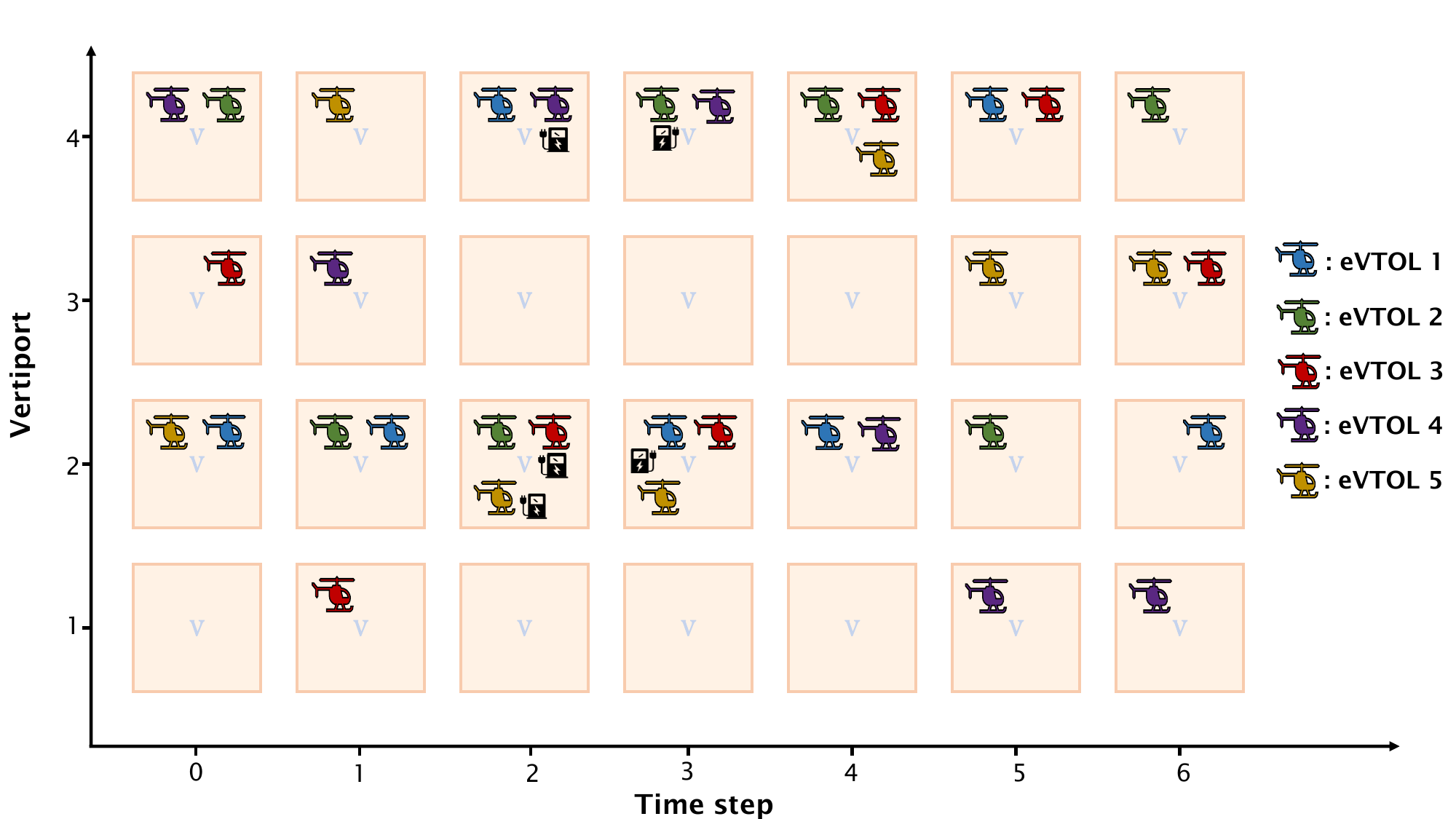}
\caption{Dispatch decisions made by multi-agent DQL eVTOL dispatch algorithm}
\label{Decisions}
\end{figure}

The dispatch decisions involved in the optimal dispatch policy learned by the DRL agents are depicted in Fig.\ref{Decisions}. At $t = 0$, the eVTOLs are in their initial positions. In successive time steps, the location of the eVTOLs get updated based on the actions taken by the agents. When the agent takes the transport action, its location changes. For example, at $t=0$, the position of eVTOL $2$ is vertiport $4$, and its position in the next time step $t = 1$ is vertiport $2$, indicating that the agent took a transport action at $t = 1$. When the positions of an eVTOL remained the same in two successive time steps, the agent either took a recharge action or a wait action in the former time step. At any given time step, the charging station icon next to an eVTOL indicates that the agent took a recharge action. The agents can be overall observed to tend to take transport actions over the longer route options available in the network, as the longer the route the higher the reward received by the agents according to Eq. \ref{equation10}. The recharge action tends to be taken by the agent either when the electricity price is low, or when an agent is nearly out of charge. If the eVTOL is not  out of charge, the agents tend to take wait action when the number of passengers in the possible routes over which it can take transport action is less than 2. This is because the operating cost break even occurs when the number of transported passengers in a trip is 2 and transporting any less result in a negative reward. The limited number of takeoff and landing pads at a vertiport influences the agents' actions. For instance, even if a longer path has enough passengers to be transported to a destination vertiport, the agent cannot choose that route if the number of takeoff and landing pads at the destination is full.

\begin{figure}[b]
\centering
\includegraphics[width=16.5cm, height = 5.5 cm]{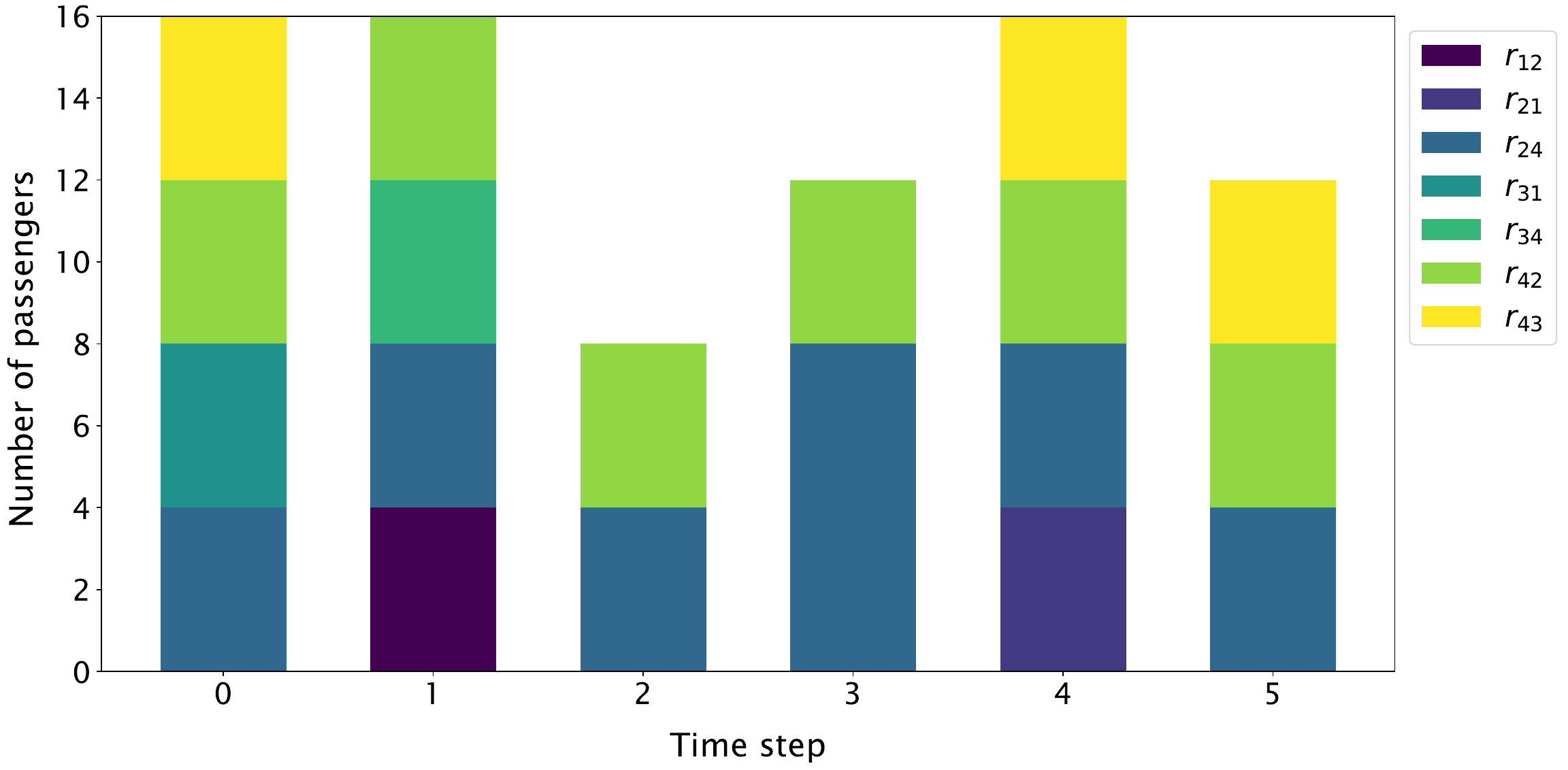}
\caption{Number of passengers transported along each route at each time step}
\label{Npassengers}
\end{figure}

Figure \ref{Npassengers} shows the number of passengers transported along each route at the different time steps. As can be seen in the figure, the number of passengers transported is higher over the the longer routes. In order of length, the routes are $r_{24}$, $r_{42}$, $r_{34}$, $r_{43}$, $r_{12}$, $r_{21}$, and $r_{31}$. If the agents did not have any passengers to transport for the longest route option available to it from its current vertiport location, it tended to select the next longest path option it had. 

\begin{figure} 
     \centering
     \begin{subfigure}[h]{0.495\textwidth}
         \centering
         \includegraphics[width=\textwidth]{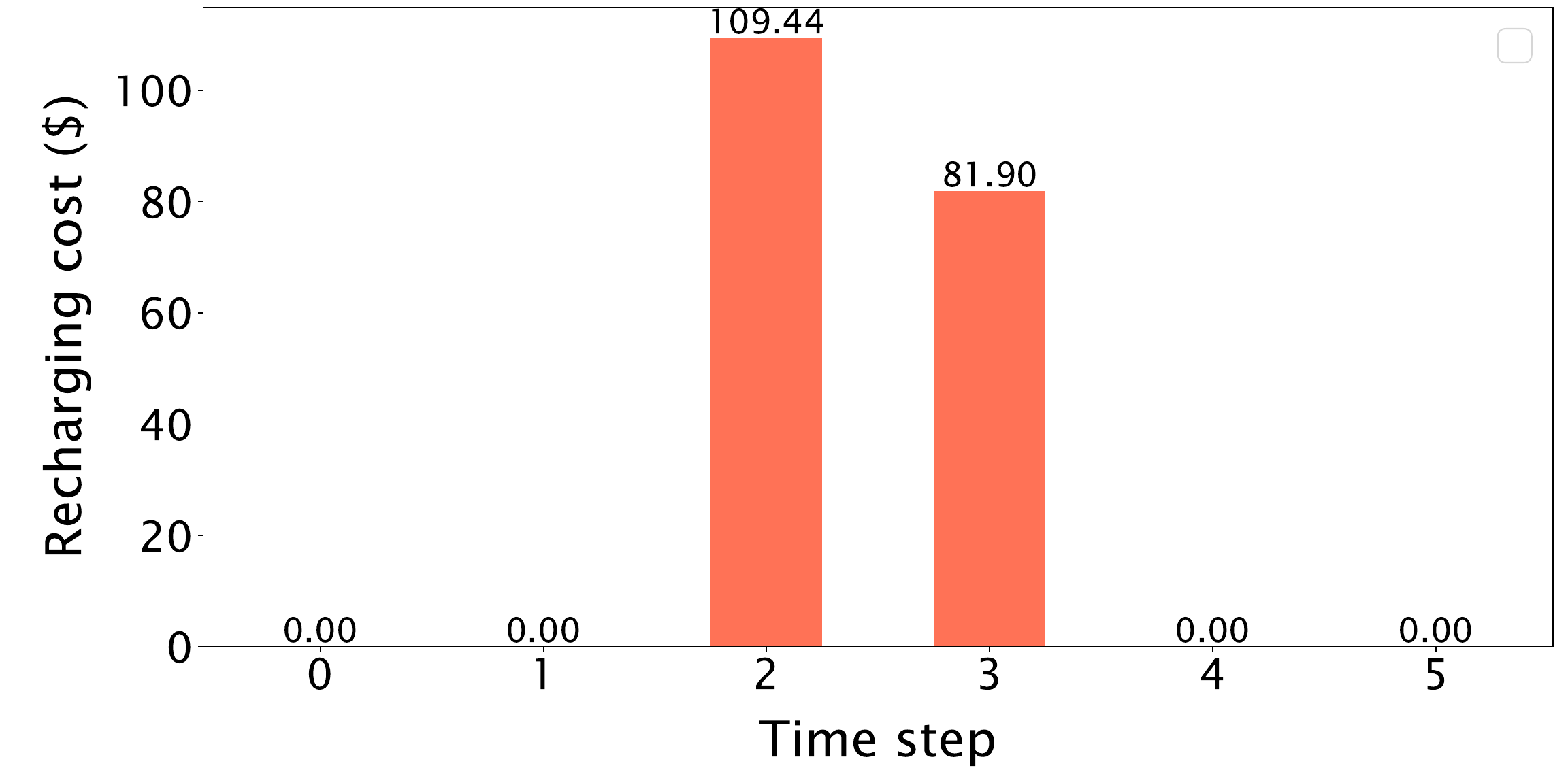}
         \caption{Recharging cost in each time step}
         \label{Recharging}
     \end{subfigure}
     \hfill
     \begin{subfigure}[h]{0.495\textwidth}
         \centering
         \includegraphics[width=\textwidth]{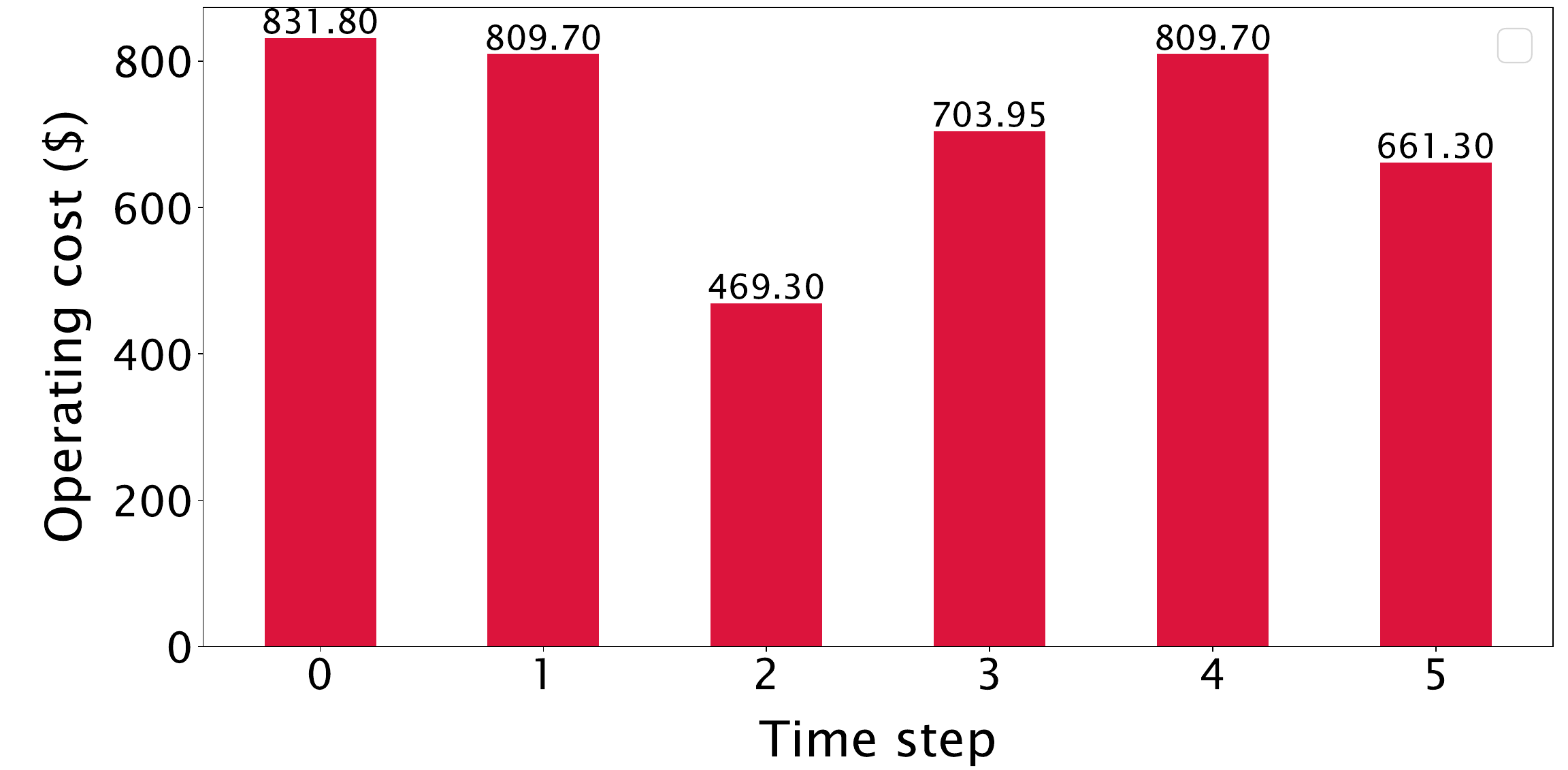}
          \caption{Operating cost in each time step}
         \label{Operating}
     \end{subfigure}
     
      \vspace{10pt}
     \centering
     \begin{subfigure}[h]{0.495\textwidth}
         \centering
         \includegraphics[width=\textwidth]{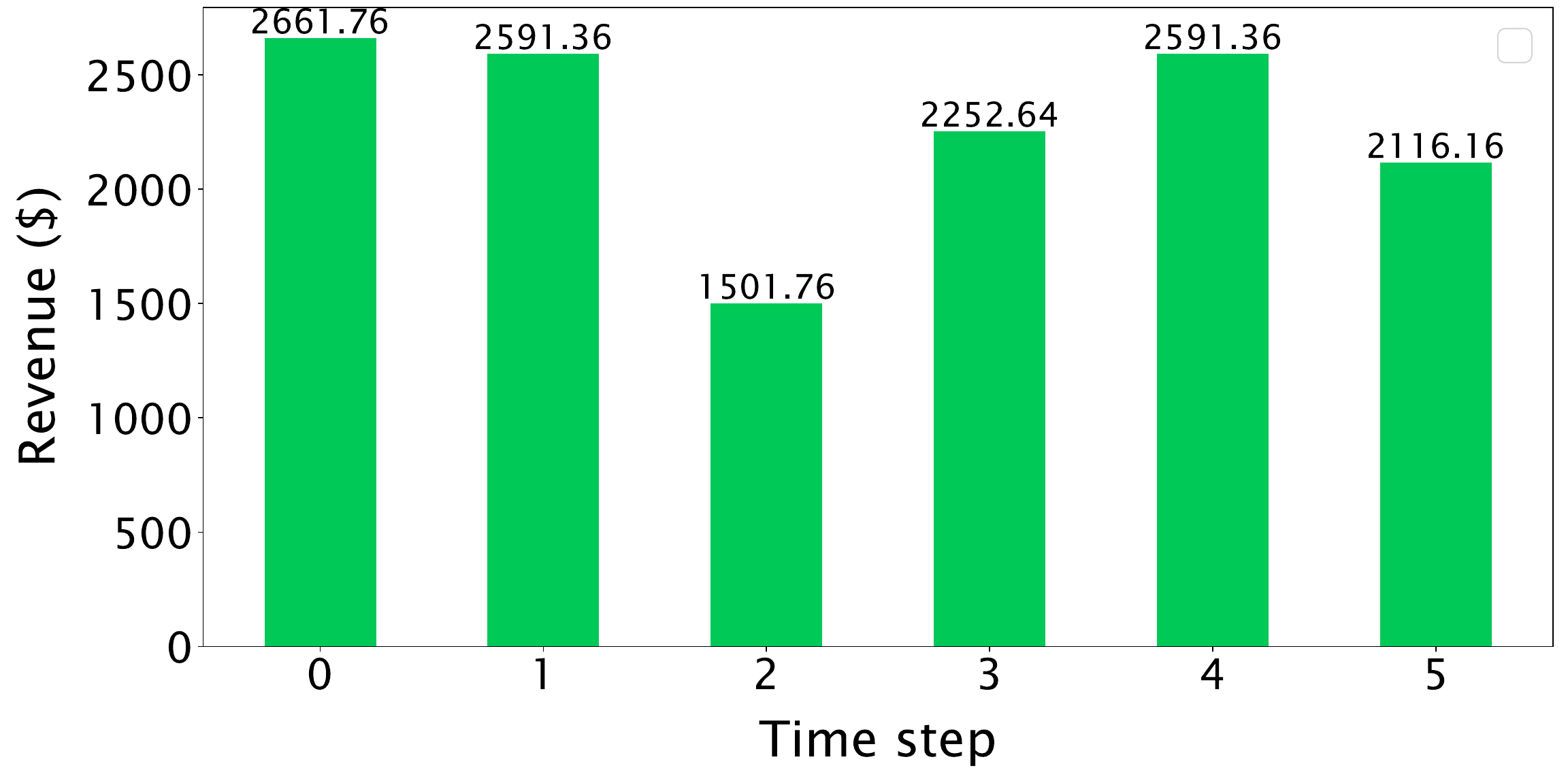}
         \caption{Revenue in each time step}
         \label{Revenue}
     \end{subfigure}
     \hfill
     \begin{subfigure}[h]{0.495\textwidth}
         \centering
         \includegraphics[width=\textwidth]{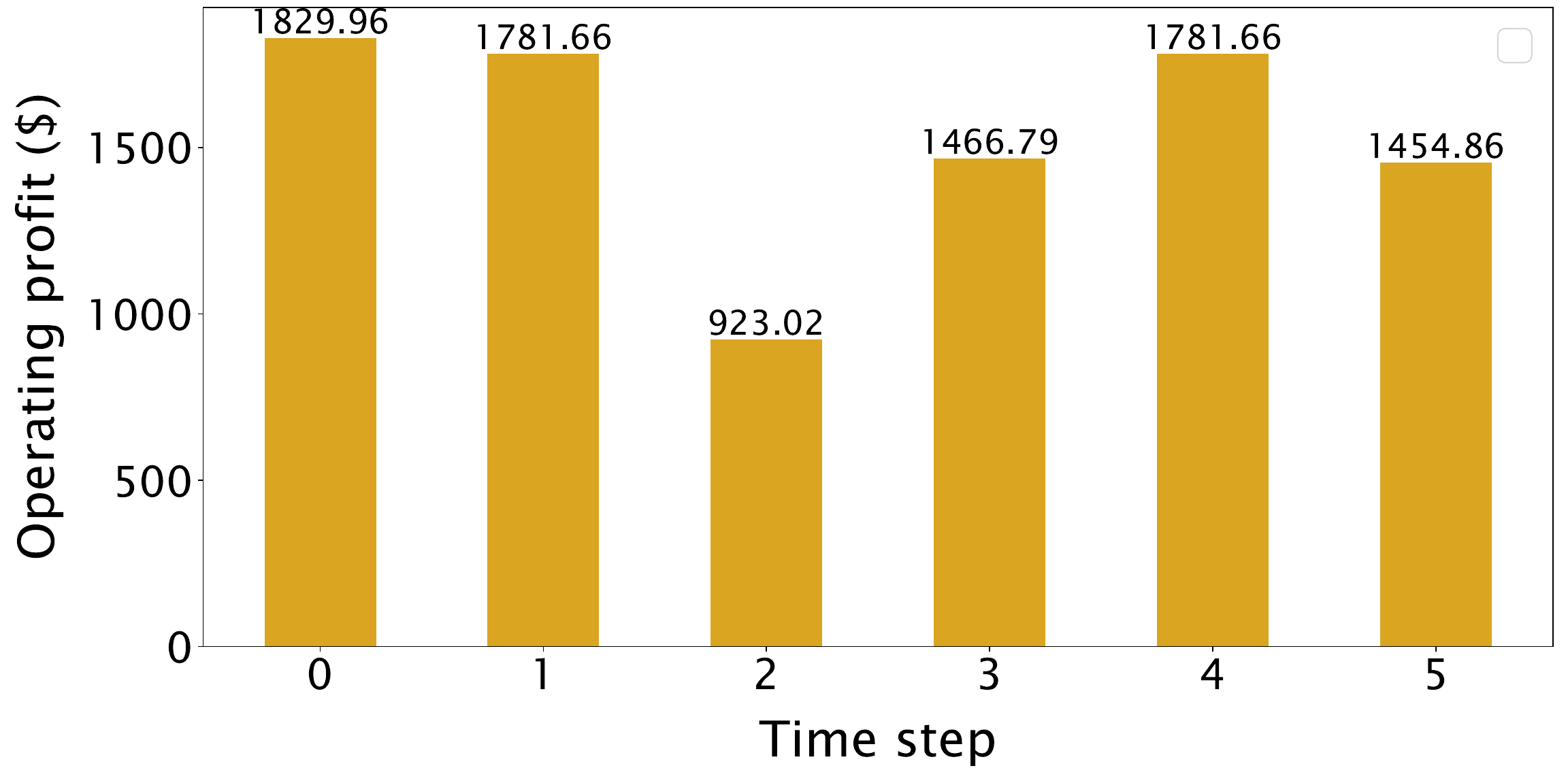}
         \caption{Operating profit in each time step}
         \label{Profit}
     \end{subfigure}
    \caption{Financial performance profile}
\label{profile}
\end{figure}

Based on the decisions made by the multi-agent DQL eVTOL dispatch algorithm, we calculate and present the revenue, operating cost, recharging cost, and operating profit in Fig. \ref{profile}. As seen in Fig. \ref{Recharging}, recharging costs are incurred only in time steps $2$ and $3$ as the agents took recharging actions only during these time steps. Specifically, three of them selected recharge actions in time step $2$, while two took the recharge action in time step $3$, as indicated in Fig. \ref{Decisions}. For all other time steps, the recharging cost is zero. The revenue, operating cost, and corresponding profit are higher in time steps 0, 1, and 4 because the agents took transportation actions over the long routes with full number of passengers.

\subsection{Single-agent Deep Q-learning eVTOL Dispatch Algorithm}
\label{RSA}
In this section, we evaluate the performance of single-agent DQL eVTOL dispatch algorithm implemented in this paper in three different numerical cases out of $36$ numerical cases. The performance of all three algorithms are compared in three different blocks when the number of eVTOLs is 5 and the number of vertiports is 3. Table \ref{tab:profitDRLC} shows the operating profit achieved by three algorithms. As demonstrated in Table \ref{tab:errorDRLC}, the performance of both the single-agent and multi-agent DQL eVTOL dispatch algorithms is approximately similar in high and medium demand blocks. However, in low demand block, the multi-agent algorithm outperforms the single-agent DQL eVTOL dispatch algorithm. As a single agent is responsible for all the eVTOLs, by increasing the number of eVTOLs, the action space gets larger exponentially, consequently, computational time for single agent DQL eVTOL dispatch algorithm during training is prohibitively long. Since the multi-agent DQL eVTOL dispatch algorithm can achieve similar or better profits to that of the single agent algorithm with much less computational time, all the numerical cases were ran for the multi-agent algorithm instead of the single agent algorithm. 

\begin{table}[!h]
\caption{Operating profit (\$) for three different algorithm ($N = 5, M = 3$)}
\label{tab:profitDRLC}
\centering
\begin{tabular}{@{}llllll@{}}
\toprule
\textbf{} & \textbf{Single-agent DQL} & \textbf{Multi-agent DQL} & \textbf{Optimization} \\\bottomrule
High demand block & \(7349.57\)  & \(7349.58\) & \(7407.49\)\\
Medium demand block & \(7407.74\)  & \(7416.32\) & \(7439.36\) \\
Low demand block & \(7312.24\)  & \(7383.4\) & \(7415.31\) \\\bottomrule
\end{tabular}
\end{table}

\begin{table}[!h]
\caption{Optimality gap of DQL algorithms ($N = 5, M = 3$)}
\label{tab:errorDRLC}
\centering
\begin{tabular}{@{}llllll@{}}
\toprule
\textbf{} & \textbf{Single-agent DQL} & \textbf{Multi-agent DQL}\\\bottomrule
High demand block  & \(0.78\%\)  & \(0.78\%\)\\
Medium demand block & \(0.43\%\)  & \(0.31\%\)\\
Low demand block & \(1.39\%\)  & \(0.43\%\)\\\bottomrule

\end{tabular}
\end{table}

\section{Conclusion}
\label{Conclusion}
Maximizing the operating profit by finding optimal eVTOL dispatch decisions is a critical objective for AAM air taxi operators. To address this, our study has implemented two DRL-based algorithms: the single-agent and multi-agent DQL eVTOL dispatch algorithms. They were designed to account for relevant operational aspects of the eVTOL dispatch problem, such as limited number of takeoff and landing pads at vertiports, limited eVTOL battery capacities, and time-varying passenger demand and electricity prices. To evaluate the performance of both algorithms, we utilized a passenger demand data set that was estimated on real-world data. Furthermore, the electricity price data used in our study corresponds to the specific vertiport locations considered in the problem. The performance of the multi-agent DQL eVTOL dispatch algorithm was benchmarked against an optimization model across $36$ different numerical cases, which featured varying number of eVTOLs, vertiports, and demand. Additionally, we conducted a comparative analysis between the single-agent and multi-agent DQL algorithms in terms of the optimality gap in operating profit achieved by each. Our findings reveal that the optimality gaps of multi-agent eVTOL dispatch algorithm, ranging between $0.23 \%$ to $4.05 \%$, were found to be consistently small, with an average of $1.75 \%$ across all numerical cases. This suggests that the multi-agent eVTOL dispatch algorithm can learn a dispatch policy which is close to the optimal dispatch policy. The optimality gap was observed to slightly increase as the scale of the problem increased. A significant difference in the solution run times were observed between the eVTOL dispatch DQL algorithms and the optimization model. The training times of the multi-agent DQL algorithm was  found to be under one hour in a standard computer with four cores. As the optimization model was not converging after one day of run time in an equivalent computer, it had to run in a supercomputer with 28 cores, where the run time was one hour on average. The comparison between the single-agent and the multi-agent eVTOL dispatch algorithms shows that the multi-agent eVTOL dispatch algorithm generates more profit than the single-agent counterpart. Furthermore, the training time for the multi-agent eVTOL dispatch algorithm was significantly shorter than that of the single-agent approach. This is because in single-agent eVTOL dispatch algorithm, the single agent is responsible for taking actions for all eVTOLs, leading to an exponential increase in the size of the action space with increase in the number of eVTOLs.

As part of future work, the dispatch problem and its DQL-based solution algorithm for eVTOL dispatch can be further developed in several ways. One approach involves incorporating dynamic trip fare and stochastic demand, travel times, and weather conditions to make the problem and algorithm more practical. Additionally, rather than assuming identical eVTOLs, the algorithm could be extended to consider a heterogeneous eVTOL fleet with different passenger capacity and battery capacity for each eVTOL. A modular eVTOL dispatch approach can also be explored to make the training process less challenging for the DRL agents. In this approach, each DRL agent would be in charge of one or more eVTOLs across only a subset of vertiports and routes within the entire vertiport network, which can reduce the size of the state-action space. Lastly, it was assumed in this work that the entire region is serviced by a single AAM air taxi operator. Multiple operators can be considered to operate in the same service area and the eVTOL dispatch algorithm can be developed to account for both competitive and cooperative agents within the same service area.

\bibliography{main}

\begin{thebibliography}{84}
\newcommand{\enquote}[1]{``#1''}
\providecommand{\natexlab}[1]{#1}
\providecommand{\url}[1]{\texttt{#1}}
\providecommand{\urlprefix}{URL }
\expandafter\ifx\csname urlstyle\endcsname\relax
  \providecommand{\doi}[1]{\discretionary{}{}{}https://doi.org/#1}\else
  \providecommand{\doi}[1]{\discretionary{}{}{}\urlstyle{rm}\url{https://doi.org/#1}}\fi

\bibitem[{USNews(2023)}]{Traffic}
USNews, \enquote{The 10 U.S. Cities with the Worst Traffic,} , 2023.
\newblock \urlprefix\url{https://www.usnews.com/news/cities/articles/10-cities-with-the-worst-traffic-in-the-us}, last accessed 23 August 2023.

\bibitem[{Northeast(2020)}]{northeast2020advanced}
Northeast, U., \enquote{Advanced Air Mobility (AAM) Vertiport Automation Trade Study,} 2020.

\bibitem[{NASA(2019)}]{NASA1}
NASA, \enquote{UAM Overview,} , 2019.
\newblock \urlprefix\url{https://www.nasa.gov/uam-overview/}, last accessed 26 June 2023.

\bibitem[{NASA(2023)}]{NASA}
NASA, \enquote{Aeronautics,} , 2023.
\newblock \urlprefix\url{https://www.nasa.gov/press-release/nasa-begins-air-taxi-flight-testing-with-joby}, last accessed 31 January 2023.

\bibitem[{Volocopter(2023)}]{Volocopter}
Volocopter, \enquote{Urban Air Mobility,} , 2023.
\newblock \urlprefix\url{https://www.volocopter.com/urban-air-mobility/}, last accessed 31 January 2023.

\bibitem[{Uber(2023)}]{Uber}
Uber, \enquote{Demonstrated the viability of a self-piloted, electric urban air mobility vehicle,} , 2023.
\newblock \urlprefix\url{https://www.uber.com/us/en/elevate/vision/}, last accessed 31 January 2023.

\bibitem[{Aitbus(2023)}]{Airbus}
Aitbus, \enquote{Demonstrated the viability of a self-piloted, electric urban air mobility vehicle,} , 2023.
\newblock \urlprefix\url{https://acubed.airbus.com/projects/vahana/}, last accessed 31 January 2023.

\bibitem[{Aurora(2023)}]{Aurora}
Aurora, \enquote{Shaping the Future of Mobility,} , 2023.
\newblock \urlprefix\url{https://www.aurora.aero/urban-air-mobility/}, last accessed 31 January 2023.

\bibitem[{Aviation(2023{\natexlab{a}})}]{JobyAviation}
Aviation, J., \enquote{Electric Aerial Ridesharing,} , 2023{\natexlab{a}}.
\newblock \urlprefix\url{https://www.jobyaviation.com/}, last accessed 31 January 2023.

\bibitem[{Goyal et~al.(2021)Goyal, Reiche, Fernando, and Cohen}]{Demand1}
Goyal, R., Reiche, C., Fernando, C., and Cohen, A., \enquote{Advanced air mobility: Demand analysis and market potential of the airport shuttle and air taxi markets,} \emph{Sustainability}, Vol.~13, No.~13, 2021, p. 7421.

\bibitem[{Rimjha(2022)}]{Demand2}
Rimjha, M., \enquote{Urban Air Mobility: Demand Estimation and Feasibility Analysis,} Ph.D. thesis, Virginia Tech, 2022.

\bibitem[{Vale~de Almeida~Norte et~al.(2023)Vale~de Almeida~Norte, Fulton, Harvey, Plevier, Oosterholt, Mendez~Garcia, Wichers, Van~Vliet, Ottens, Bootsma et~al.}]{vale2023modelling}
Vale~de Almeida~Norte, M., Fulton, M., Harvey, A., Plevier, C., Oosterholt, L., Mendez~Garcia, M., Wichers, F., Van~Vliet, L., Ottens, J., Bootsma, S., et~al., \enquote{Modelling Urban Air Mobility Demand: the Example of the {\^I}le-de-France Region,} \emph{AIAA AVIATION 2023 Forum}, 2023, p. 4105.

\bibitem[{Preis(2023)}]{preis2023estimating}
Preis, L., \enquote{Estimating vertiport passenger throughput capacity for prominent eVTOL designs,} \emph{CEAS Aeronautical Journal}, 2023, pp. 1--16.

\bibitem[{Johnson and Silva(2022)}]{aircraft1}
Johnson, W., and Silva, C., \enquote{NASA concept vehicles and the engineering of advanced air mobility aircraft,} \emph{The Aeronautical Journal}, Vol. 126, No. 1295, 2022, pp. 59--91.

\bibitem[{Bridgelall et~al.(2023)Bridgelall, Askarzadeh, and Tolliver}]{aircraft2}
Bridgelall, R., Askarzadeh, T., and Tolliver, D.~D., \enquote{Introducing an efficiency index to evaluate eVTOL designs,} \emph{Technological Forecasting and Social Change}, Vol. 191, 2023, p. 122539.

\bibitem[{Zhang et~al.(2023)Zhang, Barakos, Foster et~al.}]{zhang2023multi}
Zhang, T., Barakos, G.~N., Foster, M., et~al., \enquote{Multi-fidelity aerodynamic design and analysis of propellers for a heavy-lift eVTOL,} \emph{Aerospace Science and Technology}, Vol. 135, 2023, p. 108185.

\bibitem[{Wang et~al.(2023{\natexlab{a}})Wang, Yang, Balchanos, and Mavris}]{path1}
Wang, X., Yang, P. P.-J., Balchanos, M., and Mavris, D., \enquote{Urban Airspace Route Planning for Advanced Air Mobility Operations,} \emph{International Conference on Computers in Urban Planning and Urban Management}, Springer, 2023{\natexlab{a}}, pp. 193--211.

\bibitem[{Karpinski et~al.(2023)Karpinski, Blakesley, Krol, Anvari, Gorospe, and Sun}]{path2}
Karpinski, T., Blakesley, A., Krol, J., Anvari, B., Gorospe, G.~E., and Sun, L., \enquote{Energy-Minimization Path Planning and Control of Unmanned Aerial Systems for Advanced Air Mobility,} \emph{AIAA SCITECH 2023 Forum}, 2023, p. 0303.

\bibitem[{Wang et~al.(2021)Wang, Diepolder, Zhang, S{\"o}pper, and Holzapfel}]{traject1}
Wang, M., Diepolder, J., Zhang, S., S{\"o}pper, M., and Holzapfel, F., \enquote{Trajectory optimization-based maneuverability assessment of eVTOL aircraft,} \emph{Aerospace Science and Technology}, Vol. 117, 2021, p. 106903.

\bibitem[{Park et~al.(2023)Park, Kim, Suk, and Kim}]{traject2}
Park, J., Kim, I., Suk, J., and Kim, S., \enquote{Trajectory optimization for takeoff and landing phase of UAM considering energy and safety,} \emph{Aerospace Science and Technology}, Vol. 140, 2023, p. 108489.

\bibitem[{Xiang et~al.(2022)Xiang, Ye, Zhu, Gu, Xie, and Men}]{xiang2022multi}
Xiang, S., Ye, M., Zhu, S., Gu, J., Xie, A., and Men, Z., \enquote{A Multi-stage Precision Landing Method for Autonomous eVTOL Based on Multi-marker Joint Localization,} \emph{2022 IEEE International Conference on Robotics and Biomimetics (ROBIO)}, IEEE, 2022, pp. 1--6.

\bibitem[{Deniz et~al.(2023)Deniz, Wu, Shi, and Wang}]{deniz2023autonomous}
Deniz, S., Wu, Y., Shi, Y., and Wang, Z., \enquote{Autonomous Landing of eVTOL Vehicles via Deep Q-Networks,} \emph{AIAA AVIATION 2023 Forum}, 2023, p. 4499.

\bibitem[{Wu et~al.(2023)Wu, Deniz, Shi, Wang, and Huang}]{wu2023precision}
Wu, Y., Deniz, S., Shi, Y., Wang, Z., and Huang, D., \enquote{Precision Landing Trajectory Optimization for eVTOL Vehicles with High-Fidelity Aerodynamic Models,} \emph{AIAA AVIATION 2023 Forum}, 2023, p. 3409.

\bibitem[{Wang et~al.(2023{\natexlab{b}})Wang, Wu, and Huang}]{wang2023optimal}
Wang, Z., Wu, Y., and Huang, D., \enquote{Optimal Landing Control of eVTOL Vehicles Using ODE-Based Aerodynamic Model,} \emph{AIAA SCITECH 2023 Forum}, 2023{\natexlab{b}}, p. 1742.

\bibitem[{Li et~al.(2023)Li, Yang, Li, Zhou, and Huang}]{vertiport1}
Li, J., Yang, R., Li, C., Zhou, Y., and Huang, L., \enquote{Initial Research on The Vertiport for The Urban Air Mobility,} \emph{Proceedings of the 2nd International Conference on Information, Control and Automation, ICICA 2022, December 2-4, 2022, Chongqing, China}, 2023.

\bibitem[{Mendonca et~al.(2022)Mendonca, Murphy, Patterson, Alexander, Juarex, and Harper}]{vertiport2}
Mendonca, N., Murphy, J., Patterson, M.~D., Alexander, R., Juarex, G., and Harper, C., \enquote{Advanced Air Mobility Vertiport Considerations: A List and Overview,} \emph{AIAA AVIATION 2022 Forum}, 2022, p. 4073.

\bibitem[{Yang et~al.(2021)Yang, Liu, Ge, Rountree, and Wang}]{yang2021challenges}
Yang, X.-G., Liu, T., Ge, S., Rountree, E., and Wang, C.-Y., \enquote{Challenges and key requirements of batteries for electric vertical takeoff and landing aircraft,} \emph{Joule}, Vol.~5, No.~7, 2021, pp. 1644--1659.

\bibitem[{Zaid et~al.(2021)Zaid, Belmekki, and Alouini}]{zaid2021evtol}
Zaid, A.~A., Belmekki, B. E.~Y., and Alouini, M.-S., \enquote{eVTOL Communications and Networking in UAM: Requirements, Key Enablers, and Challenges,} \emph{arXiv preprint arXiv:2110.08830}, 2021.

\bibitem[{Bauranov and Rakas(2021)}]{review1}
Bauranov, A., and Rakas, J., \enquote{Designing airspace for urban air mobility: A review of concepts and approaches,} \emph{Progress in Aerospace Sciences}, Vol. 125, 2021, p. 100726.

\bibitem[{Afari et~al.(2023)Afari, Golubev, Lyrintzis, and Mankbadi}]{review2}
Afari, S., Golubev, V., Lyrintzis, A.~S., and Mankbadi, R., \enquote{Review of Control Technologies for Quiet Operations of Advanced Air-Mobility,} \emph{Applied Sciences}, Vol.~13, No.~4, 2023, p. 2543.

\bibitem[{Olanipekun et~al.(2023)Olanipekun, Montalvo, Salau, and Simolowo}]{review3}
Olanipekun, O.~A., Montalvo, C.~J., Salau, T.~A., and Simolowo, E.~O., \enquote{Unmanned Aerial Vehicles: A 21st Century Review of Advanced Air Mobility Platforms,} \emph{AIAA AVIATION 2023 Forum}, 2023, p. 4044.

\bibitem[{Andersson(2021)}]{mint1}
Andersson, T., \enquote{A Comparative Study on a Dynamic Pickup and Delivery Problem: Improving routing and order assignment in same-day courier operations,} , 2021.

\bibitem[{Jia et~al.(2009)Jia, Wang, and Wang}]{mint2}
Jia, Y., Wang, C., and Wang, L., \enquote{A rolling horizon procedure for dynamic pickup and delivery problem with time windows,} \emph{2009 IEEE International Conference on Automation and Logistics}, IEEE, 2009, pp. 2087--2091.

\bibitem[{Liang et~al.(2019)Liang, Zhan, Zhang, and Zhang}]{liang2019efficient}
Liang, D., Zhan, Z.-H., Zhang, Y., and Zhang, J., \enquote{An efficient ant colony system approach for new energy vehicle dispatch problem,} \emph{IEEE Transactions on Intelligent Transportation Systems}, Vol.~21, No.~11, 2019, pp. 4784--4797.

\bibitem[{B{\'e}langer et~al.(2019)B{\'e}langer, Ruiz, and Soriano}]{belanger2019recent}
B{\'e}langer, V., Ruiz, A., and Soriano, P., \enquote{Recent optimization models and trends in location, relocation, and dispatching of emergency medical vehicles,} \emph{European Journal of Operational Research}, Vol. 272, No.~1, 2019, pp. 1--23.

\bibitem[{Hu et~al.(2017)Hu, Mao, and Wei}]{energy}
Hu, W., Mao, J., and Wei, K., \enquote{Energy-efficient rail guided vehicle routing for two-sided loading/unloading automated freight handling system,} \emph{European Journal of Operational Research}, Vol. 258, No.~3, 2017, pp. 943--957.

\bibitem[{Zou et~al.(2020)Zou, Pan, Meng, Gao, and Wang}]{zou2020effective}
Zou, W.-Q., Pan, Q.-K., Meng, T., Gao, L., and Wang, Y.-L., \enquote{An effective discrete artificial bee colony algorithm for multi-AGVs dispatching problem in a matrix manufacturing workshop,} \emph{Expert Systems with Applications}, Vol. 161, 2020, p. 113675.

\bibitem[{Su et~al.(2015)Su, Luo, and Huang}]{ocost1}
Su, Q., Luo, Q., and Huang, S.~H., \enquote{Cost-effective analyses for emergency medical services deployment: A case study in Shanghai,} \emph{International Journal of Production Economics}, Vol. 163, 2015, pp. 112--123.

\bibitem[{Mes et~al.(2010)Mes, van~der Heijden, and Schuur}]{maxp1}
Mes, M., van~der Heijden, M., and Schuur, P., \enquote{Look-ahead strategies for dynamic pickup and delivery problems,} \emph{OR spectrum}, Vol.~32, 2010, pp. 395--421.

\bibitem[{Bertsimas et~al.(2019)Bertsimas, Jaillet, and Martin}]{maxp2}
Bertsimas, D., Jaillet, P., and Martin, S., \enquote{Online vehicle routing: The edge of optimization in large-scale applications,} \emph{Operations Research}, Vol.~67, No.~1, 2019, pp. 143--162.

\bibitem[{Su et~al.(2022)Su, Li, Li, and Cheng}]{maxp3}
Su, Z., Li, W., Li, J., and Cheng, B., \enquote{Heterogeneous fleet vehicle scheduling problems for dynamic pickup and delivery problem with time windows in shared logistics platform: Formulation, instances and algorithms,} \emph{International Journal of Systems Science: Operations \& Logistics}, Vol.~9, No.~2, 2022, pp. 199--223.

\bibitem[{Ghiani et~al.(2009)Ghiani, Manni, Quaranta, and Triki}]{serviclevel1}
Ghiani, G., Manni, E., Quaranta, A., and Triki, C., \enquote{Anticipatory algorithms for same-day courier dispatching,} \emph{Transportation Research Part E: Logistics and Transportation Review}, Vol.~45, No.~1, 2009, pp. 96--106.

\bibitem[{Cheng et~al.(2017)Cheng, Liao, and Hua}]{serviclevel2}
Cheng, X., Liao, S., and Hua, Z., \enquote{A policy of picking up parcels for express courier service in dynamic environments,} \emph{International Journal of Production Research}, Vol.~55, No.~9, 2017, pp. 2470--2488.

\bibitem[{Cai et~al.(2023)Cai, Zhu, Lin, Ma, Li, and Ming}]{cai2023survey}
Cai, J., Zhu, Q., Lin, Q., Ma, L., Li, J., and Ming, Z., \enquote{A Survey of Dynamic Pickup and Delivery Problems,} \emph{Neurocomputing}, 2023, p. 126631.

\bibitem[{Reyes et~al.(2018)Reyes, Erera, Savelsbergh, Sahasrabudhe, and O’Neil}]{opt1}
Reyes, D., Erera, A., Savelsbergh, M., Sahasrabudhe, S., and O’Neil, R., \enquote{The meal delivery routing problem,} \emph{Optimization Online}, Vol. 6571, 2018.

\bibitem[{Liu et~al.(2018)Liu, Tan, Kurniawan, Zhang, and Sun}]{opt2}
Liu, S., Tan, P.~H., Kurniawan, E., Zhang, P., and Sun, S., \enquote{Dynamic scheduling for pickup and delivery with time windows,} \emph{2018 IEEE 4th World Forum on Internet of Things (WF-IoT)}, IEEE, 2018, pp. 767--770.

\bibitem[{Yin et~al.(2022)Yin, Qin, and Huang}]{yin2022optimal}
Yin, W., Qin, X., and Huang, Z., \enquote{Optimal dispatching of large-scale electric vehicles into grid based on improved second-order cone,} \emph{Energy}, Vol. 254, 2022, p. 124346.

\bibitem[{Mao et~al.(2020)Mao, Liu, and Shen}]{mao2020dispatch}
Mao, C., Liu, Y., and Shen, Z.-J.~M., \enquote{Dispatch of autonomous vehicles for taxi services: A deep reinforcement learning approach,} \emph{Transportation Research Part C: Emerging Technologies}, Vol. 115, 2020, p. 102626.

\bibitem[{Mitrovi{\'c}-Mini{\'c} et~al.(2004)Mitrovi{\'c}-Mini{\'c}, Krishnamurti, and Laporte}]{heuristic1}
Mitrovi{\'c}-Mini{\'c}, S., Krishnamurti, R., and Laporte, G., \enquote{Double-horizon based heuristics for the dynamic pickup and delivery problem with time windows,} \emph{Transportation Research Part B: Methodological}, Vol.~38, No.~8, 2004, pp. 669--685.

\bibitem[{Sun et~al.(2019)Sun, Yang, Shi, and Zheng}]{heuristic2}
Sun, B., Yang, Y., Shi, J., and Zheng, L., \enquote{Dynamic pick-up and delivery optimization with multiple dynamic events in real-world environment,} \emph{IEEE Access}, Vol.~7, 2019, pp. 146209--146220.

\bibitem[{Okulewicz and Ma{\'n}dziuk(2019)}]{metaheuristic1}
Okulewicz, M., and Ma{\'n}dziuk, J., \enquote{A metaheuristic approach to solve dynamic vehicle routing problem in continuous search space,} \emph{Swarm and Evolutionary Computation}, Vol.~48, 2019, pp. 44--61.

\bibitem[{Sabar et~al.(2019)Sabar, Bhaskar, Chung, Turky, and Song}]{metaheuristic2}
Sabar, N.~R., Bhaskar, A., Chung, E., Turky, A., and Song, A., \enquote{A self-adaptive evolutionary algorithm for dynamic vehicle routing problems with traffic congestion,} \emph{Swarm and evolutionary computation}, Vol.~44, 2019, pp. 1018--1027.

\bibitem[{Trachanatzi et~al.(2020)Trachanatzi, Rigakis, Marinaki, and Marinakis}]{metaheuristic3}
Trachanatzi, D., Rigakis, M., Marinaki, M., and Marinakis, Y., \enquote{A firefly algorithm for the environmental prize-collecting vehicle routing problem,} \emph{Swarm and Evolutionary Computation}, Vol.~57, 2020, p. 100712.

\bibitem[{Sarasola et~al.(2016)Sarasola, Doerner, Schmid, and Alba}]{metaheuristic4}
Sarasola, B., Doerner, K.~F., Schmid, V., and Alba, E., \enquote{Variable neighborhood search for the stochastic and dynamic vehicle routing problem,} \emph{Annals of Operations Research}, Vol. 236, 2016, pp. 425--461.

\bibitem[{Guo and Xu(2020)}]{guo2020deep}
Guo, G., and Xu, Y., \enquote{A deep reinforcement learning approach to ride-sharing vehicle dispatching in autonomous mobility-on-demand systems,} \emph{IEEE Intelligent Transportation Systems Magazine}, Vol.~14, No.~1, 2020, pp. 128--140.

\bibitem[{Holler et~al.(2019)Holler, Vuorio, Qin, Tang, Jiao, Jin, Singh, Wang, and Ye}]{holler2019deep}
Holler, J., Vuorio, R., Qin, Z., Tang, X., Jiao, Y., Jin, T., Singh, S., Wang, C., and Ye, J., \enquote{Deep reinforcement learning for multi-driver vehicle dispatching and repositioning problem,} \emph{2019 IEEE International Conference on Data Mining (ICDM)}, IEEE, 2019, pp. 1090--1095.

\bibitem[{Liu et~al.(2022)Liu, Wu, Lyu, Li, Ye, and Qu}]{liu2022deep}
Liu, Y., Wu, F., Lyu, C., Li, S., Ye, J., and Qu, X., \enquote{Deep dispatching: A deep reinforcement learning approach for vehicle dispatching on online ride-hailing platform,} \emph{Transportation Research Part E: Logistics and Transportation Review}, Vol. 161, 2022, p. 102694.

\bibitem[{Shi et~al.(2019)Shi, Gao, Wang, Yu, and Ioannou}]{shi2019operating}
Shi, J., Gao, Y., Wang, W., Yu, N., and Ioannou, P.~A., \enquote{Operating electric vehicle fleet for ride-hailing services with reinforcement learning,} \emph{IEEE Transactions on Intelligent Transportation Systems}, Vol.~21, No.~11, 2019, pp. 4822--4834.

\bibitem[{Feng et~al.(2021)Feng, Gluzman, and Dai}]{feng2021scalable}
Feng, J., Gluzman, M., and Dai, J.~G., \enquote{Scalable deep reinforcement learning for ride-hailing,} \emph{2021 American Control Conference (ACC)}, IEEE, 2021, pp. 3743--3748.

\bibitem[{Oda and Joe-Wong(2018)}]{oda2018movi}
Oda, T., and Joe-Wong, C., \enquote{MOVI: A model-free approach to dynamic fleet management,} \emph{IEEE INFOCOM 2018-IEEE Conference on Computer Communications}, IEEE, 2018, pp. 2708--2716.

\bibitem[{Shihab et~al.(2019)Shihab, Wei, Ramirez, Mesa-Arango, and Bloebaum}]{shihab2019schedule}
Shihab, S. A.~M., Wei, P., Ramirez, D. S.~J., Mesa-Arango, R., and Bloebaum, C., \enquote{By schedule or on demand?-a hybrid operation concept for urban air mobility,} \emph{AIAA Aviation 2019 Forum}, 2019, p. 3522.

\bibitem[{Shihab et~al.(2020)Shihab, Wei, Shi, and Yu}]{shihab2020optimal}
Shihab, S. A.~M., Wei, P., Shi, J., and Yu, N., \enquote{Optimal evtol fleet dispatch for urban air mobility and power grid services,} \emph{AIAA Aviation 2020 Forum}, 2020, p. 2906.

\bibitem[{Roy et~al.(2022)Roy, Kotwicz~Herniczek, Leonard, German, and Garrow}]{roy2022flight}
Roy, S., Kotwicz~Herniczek, M.~T., Leonard, C., German, B.~J., and Garrow, L.~A., \enquote{Flight scheduling and fleet sizing for an airport shuttle air taxi service,} \emph{Journal of Air Transportation}, Vol.~30, No.~2, 2022, pp. 49--58.

\bibitem[{Sutton and Barto(2018)}]{sutton2018reinforcement}
Sutton, R.~S., and Barto, A.~G., \emph{Reinforcement learning: An introduction}, MIT press, 2018.

\bibitem[{Paul and Chowdhury(2022)}]{paul2022graph}
Paul, S., and Chowdhury, S., \enquote{A Graph-based Reinforcement Learning Framework for Urban Air Mobility Fleet Scheduling,} \emph{AIAA AVIATION 2022 Forum}, 2022, p. 3911.

\bibitem[{Mnih et~al.(2015)Mnih, Kavukcuoglu, Silver, Rusu, Veness, Bellemare, Graves, Riedmiller, Fidjeland, Ostrovski et~al.}]{mnih2015human}
Mnih, V., Kavukcuoglu, K., Silver, D., Rusu, A.~A., Veness, J., Bellemare, M.~G., Graves, A., Riedmiller, M., Fidjeland, A.~K., Ostrovski, G., et~al., \enquote{Human-level control through deep reinforcement learning,} \emph{nature}, Vol. 518, No. 7540, 2015, pp. 529--533.

\bibitem[{Mnih et~al.(2016)Mnih, Badia, Mirza, Graves, Lillicrap, Harley, Silver, and Kavukcuoglu}]{mnih2016asynchronous}
Mnih, V., Badia, A.~P., Mirza, M., Graves, A., Lillicrap, T., Harley, T., Silver, D., and Kavukcuoglu, K., \enquote{Asynchronous methods for deep reinforcement learning,} \emph{International conference on machine learning}, PMLR, 2016, pp. 1928--1937.

\bibitem[{Lian et~al.(2019)Lian, Melancon, Presta, Reevesman, Spiering, and Woodbridge}]{lian2019scalable}
Lian, X., Melancon, S., Presta, J.-R., Reevesman, A., Spiering, B., and Woodbridge, D., \enquote{Scalable real-time prediction and analysis of san francisco fire department response times,} \emph{2019 IEEE SmartWorld, Ubiquitous Intelligence \& Computing, Advanced \& Trusted Computing, Scalable Computing \& Communications, Cloud \& Big Data Computing, Internet of People and Smart City Innovation (SmartWorld/SCALCOM/UIC/ATC/CBDCom/IOP/SCI)}, IEEE, 2019, pp. 694--699.

\bibitem[{Rimjha et~al.(2021)Rimjha, Hotle, Trani, and Hinze}]{rimjha2021commuter}
Rimjha, M., Hotle, S., Trani, A., and Hinze, N., \enquote{Commuter demand estimation and feasibility assessment for Urban Air Mobility in Northern California,} \emph{Transportation Research Part A: Policy and Practice}, Vol. 148, 2021, pp. 506--524.

\bibitem[{{United States Census Bureau}(2019)}]{LODES}
{United States Census Bureau}, \enquote{Longitudinal Employer-Household Dynamics,} \url{https://lehd.ces.census.gov/data/}, 2019.
\newblock Last accessed 15 June 2023.

\bibitem[{Preis(2021)}]{preis2021quick}
Preis, L., \enquote{Quick sizing, throughput estimating and layout planning for VTOL aerodromes--a methodology for vertiport design,} \emph{AIAA Aviation 2021 Forum}, 2021, p. 2372.

\bibitem[{Roy et~al.(2021)Roy, Kotwicz~Herniczek, German, and Garrow}]{roy2021user}
Roy, S., Kotwicz~Herniczek, M.~T., German, B.~J., and Garrow, L.~A., \enquote{User base estimation methodology for a business airport shuttle air taxi service,} \emph{Journal of Air Transportation}, Vol.~29, No.~2, 2021, pp. 69--79.

\bibitem[{B{\"o}rjesson et~al.(2012)B{\"o}rjesson, Fosgerau, and Algers}]{borjesson2012income}
B{\"o}rjesson, M., Fosgerau, M., and Algers, S., \enquote{On the income elasticity of the value of travel time,} \emph{Transportation Research Part A: Policy and Practice}, Vol.~46, No.~2, 2012, pp. 368--377.

\bibitem[{{US Department of Transportation}(2023)}]{NHTS}
{US Department of Transportation}, \enquote{National Household Travel Survey,} \url{https://nhts.ornl.gov/}, 2023.
\newblock Last accessed 26 September 2023.

\bibitem[{Aviation(2023{\natexlab{b}})}]{eVTOLtype}
Aviation, A., \enquote{Introducing Midnight – Electric Vertical Takeoff and Landing Aircraft,} \url{https://www.archer.com/}, 2023{\natexlab{b}}.
\newblock Last accessed 26 September 2023.

\bibitem[{Blain(2021)}]{eVTOL2}
Blain, L., \enquote{Archer Aviation takes billion-dollar eVTOL order from United Airlines,} \url{https://newatlas.com/aircraft/archer-aviation-evtol-united/}, 2021.
\newblock Last accessed 5 September 2023.

\bibitem[{{Future Flight}(2023{\natexlab{a}})}]{eVTOL1}
{Future Flight}, \enquote{The future of advanced air mobility,} \url{https://newatlas.com/aircraft/archer-aviation-evtol-united/}, 2023{\natexlab{a}}.
\newblock Last accessed 5 September 2023.

\bibitem[{{Future Flight}(2023{\natexlab{b}})}]{cost}
{Future Flight}, \enquote{COUNTING THE COST OF URBAN AIR MOBILITY FLIGHTS,} \url{https://www.futureflight.aero/news-article/2021-11-15/counting-cost-urban-air-mobility-flights}, 2023{\natexlab{b}}.
\newblock Last accessed 26 September 2023.

\bibitem[{Gas and Company(2023)}]{PGE}
Gas, P., and Company, E., \enquote{Explore Time-of-Use Rate Plans,} , 2023.
\newblock \urlprefix\url{https://www.pge.com/en_US/residential/rate-plans/rate-plan-options/time-of-use-base-plan/time-of-use-plan.page?}, last accessed 5 June 2023.

\bibitem[{Power(2023)}]{AMP}
Power, A.~M., \enquote{EV Charging Cost on Time-of-Use (TOU) Rate Plan,} , 2023.
\newblock \urlprefix\url{https://www.alamedamp.com/349/Electric-Vehicles}, last accessed 5 June 2023.

\bibitem[{Energy(2023)}]{SVCE}
Energy, S. V.~C., \enquote{Commercial EV,} , 2023.
\newblock \urlprefix\url{https://svcleanenergy.org/}, last accessed 5 June 2023.

\bibitem[{{Keras}(2023)}]{Keras}
{Keras}, \enquote{Reinforcement Learning,} \url{https://keras.io/examples/rl/}, 2023.
\newblock Last accessed 6 November 2023.

\bibitem[{{Ray}(2023)}]{Ray}
{Ray}, \enquote{RLlib: Industry-Grade Reinforcement Learning,} \url{https://www.ray.io/}, 2023.
\newblock Last accessed 6 November 2023.

\bibitem[{{Ohio Supercomputer Center}(2023)}]{OSC}
{Ohio Supercomputer Center}, \url{https://www.osc.edu/}, 2023.
\newblock Last accessed 6 November 2023.

\end{thebibliography}

\end{document}